%% file: regret.tex
\documentclass[10pt]{article}
\pdfoutput=1
\usepackage{amssymb,amsthm,amsmath}
\usepackage{graphicx}
\usepackage{mathrsfs}
\usepackage{fullpage}

\newtheorem{theorem}{Theorem}
\newtheorem{lemma}[theorem]{Lemma}
\newtheorem{corollary}[theorem]{Corollary}
\newtheorem{definition}[theorem]{Definition}
\theoremstyle{definition}

\include{macros}

\begin{document}

\title{A Stochastic View of Optimal Regret through Minimax Duality}

\author{
Jacob Abernethy \\
Computer Science Division \\
UC Berkeley \\
\and
Alekh Agarwal \\
Computer Science Division \\
UC Berkeley\\
\vspace{2mm}
\and
Peter L. Bartlett \\
Computer Science Division \\
Department of Statistics \\
UC Berkeley
\and 
Alexander Rakhlin \\
Department of Statistics\\
University of Pennsylvania
}

\maketitle

\begin{abstract}
	We study the regret of optimal strategies for online convex
	optimization games.  Using von Neumann's minimax theorem, we
	show that the optimal regret in this adversarial setting is
	closely related to the behavior of the empirical
	minimization algorithm in a stochastic process setting:
	it is equal to the maximum, over joint distributions of the
	adversary's action sequence, of the difference between a sum
	of minimal expected losses and the minimal empirical loss.
	We show that the optimal regret has a natural geometric
	interpretation, since it can be viewed as the gap in
	Jensen's inequality for a concave functional---the minimizer
	over the player's actions of expected loss---defined on a
	set of probability distributions.  We use this expression to
	obtain upper and lower bounds on the regret of an optimal
	strategy for a variety of online learning problems. Our
	method provides upper bounds without the need to construct a
	learning algorithm; the lower bounds provide explicit
	optimal strategies for the adversary.
\end{abstract}

\section{Introduction}
\label{sec:intro}

Within the Theory of Learning, two particular topics have gained significant popularity over the past 20 years: Statistical Learning and Online Adversarial Learning. Papers on the former typically study generalization bounds, convergence rates, complexity measures of function classes---all under the assumption that the examples are drawn, typically in an i.i.d. manner, from some underlying distribution. Working under such an assumption, Statistical Learning finds its roots in statistics, probability theory, high-dimensional geometry, and one can argue that the main questions are by now relatively well-understood.

Online Learning, while having its origins in the early 90's, recently became a popular area of research once again. One might argue that it is the assumptions, or lack thereof, that make online learning attractive. Indeed, it is often assumed that the observed data is generated maliciously rather than being drawn from some fixed distribution. Moreover, in contrast with the ``batch learning'' flavor of Statistical Learning, the sequential nature of the online problem lets the adversary change its strategy in the middle of the interaction. It is no surprise that this adversarial learning seems quite a bit more difficult than its statistical cousin. The worst case adversarial analysis does provide a realistic modeling in learning scenarios such as network security applications, email spam detection, network routing etc., which is largely responsible for the renewed interest in this area.

Upon a review of the central results in adversarial online
learning---most of which can be found in the recent book Cesa-Bianchi
and Lugosi \cite{CesaBianchiLugosi06book}---one cannot help but notice
frequent similarities between the guarantees on performance of online
algorithms and the analogous guarantees under stochastic assumptions.
However, discerning an explicit link has remained elusive.
Vovk \cite{Vovk98} notices this phenomenon: ``for some important problems, the adversarial bounds of on-line competitive learning theory are only a tiny amount worse than the average-case bounds for some stochastic strategies of Nature.''

In this paper, we attempt to build a bridge between
adversarial online learning and statistical learning.  Using
von Neumann's minimax theorem, we show that the optimal
regret of an algorithm for online convex optimization is
exactly the difference between a sum of minimal expected
losses and the minimal empirical loss, under an adversarial
choice of a stochastic process generating the data.  This leads to
upper and lower bounds for the optimal regret that exhibit several
similarities to results from statistical learning.

The online convex optimization game proceeds in rounds. At each
of these $T$ rounds, the player (learner) predicts a vector
in some convex set, and the adversary responds with a convex
function which determines the player's loss at the chosen
point. In order to emphasize the relationship with the
stochastic setting, we denote the player's choice as
$f\in\F$ and the adversary's choice as $z\in\Z$.  Note that
this differs, for instance, from the notation
in \cite{AbeBarRakTew08colt}.

Suppose $\F$ is a \emph{convex compact} class of functions,
which constitutes the set of Player's choices. The Adversary draws his choices from a
\emph{closed compact} set $\Z$. We also define a continuous bounded
loss function $\ell: \Z\times\F \to \R$ and assume that
$\ell$ is convex in the second argument. Denote by
$\ell(\F)=\{\ell(\cdot, f): f\in\F\}$ the associated loss
class. Let $\Probs$ be the set of all probability
distributions on $\Z$. Denote a sequence $(Z_1,\ldots,Z_T)$ by
$Z_1^T$. We denote a joint distribution on $\Z^T$ by a
bold-face $\jp$ and its conditional  and marginal
distributions by $\pt$ and $\ptm$, respectively.

The online convex optimization interaction is described as follows.

\frameitblack{
{\bf Online Convex Optimization (OCO) Game }
\begin{itemize}
	\setlength{\itemsep}{-1pt}
	\item[] \hspace{-0.2in} At each time step $t=1$ to $T$,
	\item Player chooses $f_t \in \F$
	\item Adversary chooses $z_t \in \Z$
	\item Player observes $z_t$ and suffers loss $\ell (z_t, f_t)$ 
\end{itemize}
}

The objective of the player is to minimize the {regret}
$$ \sum_{t=1}^T \ell(z_t, f_t) - \inf_{f\in \F} \sum_{t=1}^T \ell (z_t, f).$$

It turns out that many online learning scenarios
can be realized as instances of OCO, including
prediction with expert advice, data compression, 
sequential investment, and forecasting with side information (see, for
example,~\cite{CesaBianchiLugosi06book}). 

\section{Applying von Neumann's minimax theorem}
\label{sec:vonneuman}

Define the value of the OCO game---which we also call the {\em minimax
regret}---as
\begin{align}  
	\label{eq:def_minimax_regret}
	\Reg_T  = \inf_{f_1\in \F}\sup_{z_1\in\Z} \cdots \inf_{f_{T-1}\in \F}\sup_{z_{T-1}\in\Z} \inf_{f_T\in \F}\sup_{z_T\in\Z} \left(\sum_{t=1}^T\ell(z_t,f_t) - \inf_{f\in \F}\sum_{t=1}^T \ell(z_t,f)\right).
\end{align}
The OCO game has a purely ``optimization'' flavor. However,
applying von Neumann's minimax theorem shows that
its value is closely related to the behavior of the empirical
minimization algorithm in a stochastic process setting.

\begin{theorem}\label{theorem:maximin}
	Under the assumptions on $\F$, $\Z$, and $\ell$ given in the previous section, 
	\begin{align}
		\label{eq:regret_equality}   
		\Reg_T =  \sup_{\jp} \E \left[
		  \sum_{t=1}^T \inf_{f_t\in \F}
		  \E \left[ \ell(Z_t,f_t)|Z_1^{t-1}\right] - \inf_{f\in \F} \sum_{t=1}^T \ell(Z_t,f) \right],
	\end{align}	
	where the supremum is over all joint distributions $\jp$ on
	$\Z^T$ and the expectations are over the sequence of random variables $\{Z_1, \ldots, Z_T\}$ drawn according to $\jp$.
\end{theorem}

The proof relies on the following version of von 
Neumann's minimax theorem; it appears as Theorem~7.1
in~\cite{CesaBianchiLugosi06book}. 
\begin{proposition}
	\label{prop:vonneumann}
	Let $M(x,y)$ denote a bounded real-valued function on $\cal X \times \cal Y$, where $\cal X$ and $\cal Y$ are convex sets and $\cal X$ is compact. Suppose that $M(\cdot, y)$ is convex and continuous for each fixed $y\in \cal Y$ and $M(x, \cdot)$ is concave for each $x\in \cal X$. Then
	$$ \inf_{x\in \cal X} \sup_{y\in \cal Y} M(x,y) = \sup_{y\in \cal Y} \inf_{x\in \cal X} M(x, y).$$
\end{proposition}

\begin{proof}[{\bf Proof of Theorem~\ref{theorem:maximin}}]

For the sake of clarity, we prove the Theorem for $T=2$. The proof for $T>2$ uses essentially the same steps while the notation is less transparent. We postpone the general proof to the Appendix. 

We have
\begin{align*}  
	\Reg_2  = \inf_{f_1\in \F}\sup_{z_1\in\Z}\inf_{f_2 \in \F}\sup_{z_2\in\Z} \left(\sum_{t=1}^2\ell(z_t,f_t) - \inf_{f\in \F}\sum_{t=1}^2 \ell(z_t,f)\right).
\end{align*}

Consider the last optimization choice $z_2$. Suppose we instead draw $z_2$ according to a distribution, and compute the expected value of the quantity in the parentheses. Then it is clear that maximizing this expected value over all distributions on $\Z$ is equivalent to maximizing over $z_2$, with the optimizing distribution concentrated on the optimal point. Hence,
\begin{align}  
	\label{eq:regret_start}
    \Reg_2 = \inf_{f_1\in \F}\sup_{z_1\in\Z}\inf_{f_2 \in \F}\sup_{p_2 \in \Probs} \E_{z_2\sim p_2}\left[ \sum_{t=1}^2\ell(z_t,f_t) - \inf_{f\in \F}\sum_{t=1}^2 \ell(z_t,f)\right]. 
\end{align}

We now apply Proposition \ref{prop:vonneumann} to the last $\inf/\sup$ pair in \eqref{eq:regret_start} with 
$$M(f_2,p_2) = \E_{z_2\sim p_2}\left[ \sum_{t=1}^2 \ell(z_t, f_t) - \inf_{f\in\F}\sum_{t=1}^2 \ell(z_t, f)\right],$$
which is convex in $f_2$ (by assumption) and linear in $p_2$. Moreover, the set $\F$ is compact, and both $\F$ and $\Probs$ are convex. We conclude that
\begin{align*}  
    \Reg_2 &= \inf_{f_1\in \F}\sup_{z_1\in\Z}\inf_{f_2 \in \F}\sup_{p_2 \in \Probs} \E_{z_2\sim p_2}\left[ \sum_{t=1}^2\ell(z_t,f_t) - \inf_{f\in \F}\sum_{t=1}^2 \ell(z_t,f)\right] \\
    &= \inf_{f_1\in \F}\sup_{z_1\in\Z}\sup_{p_2 \in \Probs} \inf_{f_2 \in \F} \E_{z_2\sim p_2}\left[ \sum_{t=1}^2\ell(z_t,f_t) - \inf_{f\in \F}\sum_{t=1}^2 \ell(z_t,f)\right] \\
	&= \inf_{f_1\in \F}\sup_{z_1\in\Z}\sup_{p_2 \in \Probs} \left[\ell(z_1,f_1) + \inf_{f_2 \in \F} \E_{z\sim p_2} \ell(z, f_2) - \E_{z_2\sim p_2}\inf_{f\in \F}\sum_{t=1}^2 \ell(z_t,f) \right]\\
	&= \inf_{f_1\in \F}\sup_{z_1\in\Z} 
		\underbrace{
		\left[
			\ell(z_1,f_1) + \sup_{p_2 \in \Probs}
		 		\left\{ 
					\inf_{f_2 \in \F} \E_{z\sim p_2} \ell(z, f_2) - \E_{z_2\sim p_2}\inf_{f\in \F}\sum_{t=1}^2 \ell(z_t,f) 
				\right\}	
		\right]
		}_{A(z_1, f_1)}
\end{align*}
Now consider the supremum over $z_1$. Using the same argument as before, we have
\begin{align*}  
    \Reg_2 &= \inf_{f_1\in \F}\sup_{z_1\in\Z} A(z_1, f_1)
	=\inf_{f_1\in \F}\sup_{p_1 \in \Probs}\E_{z_1\sim p_1} A(z_1, f_1).
\end{align*}
Observe that the function
$$M(f_1,p_1) = \E_{z_1\sim p_1}  A(z_1, f_1)$$
is convex in $f_1$ and linear in $p_1$. Appealing to Proposition \ref{prop:vonneumann} again, we obtain
\begin{align*}  
    \Reg_2 &=\inf_{f_1\in \F}\sup_{p_1 \in \Probs}\E_{z_1\sim p_1} A(z_1, f_1)\\
	&= \sup_{p_1 \in \Probs}\inf_{f_1\in \F}\E_{z_1\sim p_1}  	
		\left[
			\ell(z_1,f_1) + \sup_{p_2 \in \Probs}
				\left\{
					\inf_{f_2 \in \F} \E_{z\sim p_2} \ell(z, f_2) - \E_{z_2\sim p_2}\inf_{f\in \F}\sum_{t=1}^2 \ell(z_t,f) 
				\right\}
		\right]\\
	&= \sup_{p_1 \in \Probs}
		\left[
			\left(\inf_{f_1\in \F}\E_{z\sim p_1} \ell(z,f_1)\right) 
			+ \E_{z_1\sim p_1} \sup_{p_2 \in \Probs}
				\left\{
					\inf_{f_2 \in \F} \E_{z\sim p_2} \ell(z, f_2) - \E_{z_2\sim p_2}\inf_{f\in \F}\sum_{t=1}^2 \ell(z_t,f) 
				\right\}
		\right]\\
	&= \sup_{p_1 \in \Probs}
			\E_{z_1\sim p_1} \sup_{p_2 \in \Probs}
				\left\{
					\left(\inf_{f_1\in \F}\E_{z\sim p_1} \ell(z,f_1)\right) + \inf_{f_2 \in \F} \E_{z\sim p_2} \ell(z, f_2) - \E_{z_2\sim p_2}\inf_{f\in \F}\sum_{t=1}^2 \ell(z_t,f) 
				\right\}
\end{align*}
A key observation that makes the above argument valid is that the choice of $f_1$  does not depend on $p_2$, and by the same token $p_2$ is not influenced by a particular choice of $f_1$. Now, it is easy to see that maximizing over $p_1$, then averaging over $z_1\sim p_1$, and then maximizing over $p_2(\cdot|z_1)$ is the same as maximizing over joint distributions $\jp$ on $(z_1,z_2)$ and averaging over $z_1$. Thus,
\begin{align*}  
    \Reg_2 &= 
		\sup_{\jp}
			\E_{z_1\sim p_1} 
				\left\{
					\left(\inf_{f_1\in \F}\E_{z\sim p_1} \ell(z,f_1)\right) + \left( \inf_{f_2 \in \F} \E_{z\sim p_2} \ell(z, f_2) \right) - \E_{z_2\sim p_2}\inf_{f\in \F}\sum_{t=1}^2 \ell(z_t,f) 
				\right\}
\end{align*}
which proves the Theorem for $T=2$.

\end{proof}

We can think of Eq. \eqref{eq:regret_equality} as a game where the
adversary goes first. At every round he ``plays'' a distribution and
the player responds with a function that minimizes the conditional
expectation. 

We remark that we can allow the player to choose $f_t$'s
non-deterministically in the original OCO game. In that case, the
original infimum should be over distributions on $\F$. We then do not
need convexity of $\ell$ in $f\in \F$'s in order to apply von
Neumann's theorem, and the resulting expression for the value of the
game is the same.


\section{First Steps}

The present work focuses on analyzing the expression in Equation
\eqref{eq:regret_equality} for a range of different choices of $\Z$
and $\F$, as well as for various assumptions made about the loss
function $\ell$. We are not only interested in upper- and
lower-bounding the value of the game $\Reg_T$, but also in determining
the types of distributions $\jp$ that maximize or almost maximize the
expression in \eqref{eq:regret_equality}. To that end, define
$\jp$-regret as
\begin{align}
	\label{eq:stochastic}
	\E \left[ \sum_{t=1}^T \min_{f_t\in \F} \E \left[ \ell(Z_t,f_t)|Z_1^{t-1}\right] - \min_{f\in \F} \sum_{t=1}^T \ell(Z_t,f) \right] 
\end{align}
for any joint distribution $\jp$ of $(z_1,\ldots,z_T) \in \Z^T$.
In this section we will provide an array of analytical tools for working with $\Reg_T (\jp)$.

\subsection{Regret for IID and Product Distributions} 
\label{sub:iid_product}

Let us start with a simple example. Suppose $\Z=[0,1]$, $\F = [0,1]$, and $\ell(z, f) = |z-f|$. For this game, we might try various strategies $\jp$ and compute the value $\Reg_T(\jp)$. One choice of the joint distribution $\jp$ could be to put mass on disjoint intervals of $\Z$ at each round. Suppose the conditional distribution at time $t$, $\pt$ is uniform on $\left[\frac{t-1}{T}, \frac{t}{T}\right]$. Then,  
$$f_t^* =  \arg\min_{f_t\in \F} \E \left[ \ell(Z,f_t)|Z_1^{t-1}\right] = \frac{t+1/2}{T},$$ 
the midpoint of the interval, while the minimizer over the data $\hat{f} = \arg\min_{f\in \F} \sum_{t=1}^T \ell(Z_t,f)$ is close to $\frac{1}{2}$. It is easy to check that $\Reg_T(\jp)$ is negative and linear in $T$. 

This example suggests that the chosen distribution $\jp$ is not optimal, as it forces the best decision in hindsight to be bad as compared to the intermediate decisions. The root of the problem appears in the disjoint nature of the support of the distributions. One might wonder whether this suggests that the optimal distribution should, in fact, be the same between rounds. Hence, a natural next step is to consider i.i.d. as well as general product distributions $\jp$ as candidates for maximizing $\Reg_T(\jp)$.

The following Lemma states that $\jp$-regret is non-negative for any choice of an i.i.d. distribution.

\begin{lemma}
	\label{lem:positivity}
For any i.i.d. distribution $\jp$, $\Reg_T(\jp)\geq 0$. Hence, $ \Reg_T \geq 0.$
\end{lemma}
\begin{proof}
For an i.i.d. distribution Eq. \eqref{eq:stochastic} becomes
\begin{align}
	\label{eq:iid_positive}
	\Tinv\Reg_T (\jp) &= \Tinv\sum_{t=1}^T \min_{f_t\in \F} \E \left[ \ell(Z,f_t)\right] - \E \min_{f\in \F} \Tinv\sum_{t=1}^T \ell(Z_t,f) \nonumber\\
	&= \min_{f\in \F} \E \left[ \ell(Z,f)\right] - \E \min_{f\in \F} \Tinv\sum_{t=1}^T \ell(Z_t,f) \nonumber\\
	&\geq \min_{f\in \F} \E \left[ \ell(Z,f)\right] -  \min_{f\in \F} \E \Tinv\sum_{t=1}^T \ell(Z_t,f) \nonumber\\
	&= 0 \nonumber
\end{align}
where the inequality is due to the fact that $\E\min \leq \min\E$.
\end{proof}

Observe that $\Reg_T(\jp)$ for an i.i.d.\ process is the difference
between the minimum expected loss and the expectation of the empirical
loss of an empirical minimizer.

With the goal of studying various types of distributions, we now define the following hierarchy:
\begin{align*}  
	\Reg_T^{\text{i.i.d}} := \sup_{\jp = p \times \ldots \times p} \Reg_T(\jp) &; &
	\Reg_T^{\text{indep.}} := \sup_{\jp = p_1 \times \ldots \times p_T} \Reg_T(\jp),
\end{align*}
where $p,p_1,\ldots, p_T$ are arbitrary distributions on $\Z$. It is immediately clear that
\begin{equation}
	\label{eq:hierarchy}
	0\quad \leq \quad \Reg_T^{\text{i.i.d}} \quad \leq \quad \Reg_T^{\text{indep.}} \quad \leq \quad \Reg_T.
\end{equation}

We will see that, given particular assumptions on $\F,\Z$ and $\ell$,
some of the gaps in the above hierarchy are significant, while others
are not. Before continuing, however, we need to develop some tools for
analyzing the minimax regret.


\subsection{Tools for a General Analysis} 
\label{sub:tools_for_a_general_analysis}

We now introduce two new objects that help to simplify the expression in \eqref{eq:regret_equality} as well as derive properties of $\Reg_T(\jp)$. 
\begin{definition}
	Given sets $\F,\Z$, we can define the  {\em minimum expected loss} functional $\Phi$ as
	\begin{align*}
		\Phi (p) := \inf_{f\in \F} \E_{Z\sim p} \left[\ell(Z,f)\right] 
	\end{align*}
	where $p$ is some distribution on $\Z$. 
\end{definition}
Defining an inner product $\langle h, p \rangle = \int_z h(z) dp(z)$ for a distribution $p$, we observe that $\Phi (p) = \inf_{f\in \F} \langle \ell(\cdot ,f), p \rangle$.

\begin{definition}
	For any $Z_1, \ldots, Z_T \in \Z^T$, we denote $\Unif = \Tinv\sum_{t=1}^T \one_{Z_t}(\cdot)$, the empirical distribution. 
\end{definition}

With this additional notation, we can rewrite \eqref{eq:stochastic} as
\begin{align}
	\label{eq:decomposition}
	\Tinv \Reg_T (\jp) &= \Tinv \sum_{t=1}^T \E\Phi (\pt) - \E \Phi ( \Unif).
\end{align}
Thus, the Adversary's task is to induce a large deviation between the
average sequence of conditional distributions $\{\pt\}$ and an
empirical sample $\Unif$ from these conditionals, where the deviation
is defined by way of the functional $\Phi$.
 It is easy to check that $\Phi$ and $\Reg_T$ are concave, as the next lemma shows. The proof is postponed to the Appendix.
\begin{lemma}
	\label{lem:regret_concave}
	The functional $\Phi(\cdot)$ is concave on the space of distributions over $\Z$ and $\Reg_T(\cdot)$ is concave with respect to joint distributions on $\Z^T$.
\end{lemma}

It is indeed concavity of $\Phi$ that is key to understanding the
behavior of $\Reg_T$. A hint of this can already be seen in the proof of Lemma~\ref{lem:positivity}, where the only inequality is due to the concavity of the $\min$. In the next section, we show how this description of regret can be interpreted through a \emph{Bregman divergence} in terms of $\Phi$.


\subsection{Divergences and the Gap in Jensen's Inequality} 
\label{sub:divergences_and_the_gap_in_jensen_s_inequality}

We now show how to interpret regret through the lens of Jensen's Inequality by providing yet another expression for it, now in terms of Bregman Divergences. We begin by revisiting the i.i.d. case $\jp = p^T = p \times \ldots \times p$, for some distribution $p$ on $\Z$. Equation \eqref{eq:decomposition} simplifies to a very natural quantity,
\begin{equation}\label{eq:iidphiregret}
	\Tinv \Reg_T(p^T) = \Phi(p) - \E \Phi(\Unif).
\end{equation}

Notice that $\Unif$ is a random quantity, and in particular that
$\E \Unif = p$. As $\Phi(\cdot)$ is concave, with an immediate application of Jensen's Inequality we obtain $\Reg_T(p^T)\geq 0$.
For arbitrary joint distributions $p$, we can similarly interpret regret as a ``gap'' in Jensen's Inequality, albeit with some added complexity. 
\begin{definition}
	If $F$ is any convex differentiable\footnote{Here, we mean differentiable with respect to the Fr\'echet or G\^ateaux derivative. We refer the reader to \cite{FriSriGup06functionalBregman} for precise definitions of functional Bregman Divergences.} functional on the space of distributions on $\Z$, we define {\em Bregman divergence} with respect to $F$ as
$$ \Dvg_{F} (q, p) = F(q) - F(p) - \langle \nabla F (p),  q-p \rangle.$$
\end{definition}
If $F$ is non-differentiable, we can take a particular subgradient $v_p \in \partial F(p)$ in place of $\nabla F(p)$. Note that the notion of subgradients is well-defined even for infinite-dimensional convex functions. Having chosen\footnote{The assumption of compactness of $\F$, together with the characterization of the subgradient set in Section~\ref{sec:phidiff}, allow us, for instance, to define the mapping $p\mapsto v_p$ by putting a uniform measure on the subgradient set and defining $v_p$ to be the expected subgradient with respect to it. In fact, the choice of the mapping is not important, as long as it does not depend on $q$.} a mapping $p \mapsto v_p\in \partial F(p)$, we define {\em a generalized divergence with respect to $F$ and $v_p$} as
$$ \Dvg_{F} (q, p) = F(q) - F(p) - \langle v_p, q-p \rangle.$$

Throughout the paper, we focus only on the divergence $\Dvg_{-\Phi}$, and thus we omit ${-\Phi}$ from the notation for simplicity.

Given the definition of divergence, it immediately follows that, for a random distribution $q$,
	$$ \Phi(\E q) - \E\Phi(q) = \E \Dvg (q, \E q)$$
since the linear term disappears under the expectation. This simple observation is quite useful; notice we now have an even simpler expression for i.i.d. regret \eqref{eq:iidphiregret}:
\begin{equation*}
	\Tinv \Reg (p^T) = \E \Dvg (\Unif, p).
\end{equation*}
In other words, the $p^T$-regret is {\em equal} to the expected divergence between the empirical  distribution and its expectation. This will be a starting point for obtaining lower bounds for $\Reg_T$. 
For general joint distributions $p$, let us rewrite the expression in \eqref{eq:decomposition} as 
$$\E_{t\sim U} \E\Phi (\pt) - \E \Phi ( \Unif),$$ where we replaced the average with a uniform distribution on the rounds.  Roughly speaking, the next lemma says that one can obtain $\E \Phi ( \Unif)$ from $\E_{t\sim U} \E\Phi (\pt)$ through three applications of Jensen's inequality, due to various expectations being ``pulled'' inside or outside of $\Phi$.

\begin{lemma}
Suppose $p$ is an arbitrary joint distribution. Denote by $\pt$ and $p_t^m$ the conditional and marginal distributions, respectively. Then
\begin{align}
		\label{eq:general_equality}
		\Tinv\Reg (\jp) = -\Delta_0 -\Delta_1 + \Delta_2,
\end{align}
where
$$\Delta_0 = \Tinv \sum_t \Dvg\left( \ptm , {\textstyle \Tinv \sum_{t'} \ptmp }\right), \quad \quad \Delta_1 = \Tinv \sum \E_{\jp} \Dvg (\pt, \ptm ),$$
$$\Delta_2 = \E_{\jp} \Dvg \left(\Unif, \Tinv\sum \ptm \right),$$
and $p_t^m = \E[\pt]$ is the marginal distribution at time $t$. 
\end{lemma}

\begin{proof}
	The marginal distribution satisfies
	$\E \pt = \ptm$, and it is easy to see that $\E \Unif = \Tinv \sum_t \ptm$. Given this, we see that
	\begin{align*}  
		\Tinv \Reg (\jp) &= \E_{\jp} \left[ \Tinv \sum_{t=1}^T \Phi (\pt) - \Phi (\Unif) \right] \\
		&=  \overbrace{\E_{\jp} \left[\Tinv \sum_t \left\{\Phi
		(\pt) -  \Phi(\ptm)\right\} \right]}^{-\Delta_1} \\*
		& \qquad {}
		+ \overbrace{\Tinv \sum_t \Phi(\ptm) -
		\Phi\left({\textstyle \Tinv \sum_{t}
		\ptm}\right)}^{-\Delta_0} \\*
		& \qquad {}
		- \overbrace{\E_{\jp} \left[  \Phi (\Unif) -
		  \Phi\left({\textstyle \Tinv \sum_{t} \ptm}\right)
		  \right]}^{-\Delta_2}.
	\end{align*}
\end{proof}

This lemma sheds some light on the influence of an i.i.d.\ vs.
product vs. arbitrary joint distribution on the regret. For
product distributions, every conditional distribution is identical to
its marginal distributions, thus implying $\Delta_1 = 0$. Furthermore,
for any i.i.d.\ distribution, each marginal distribution is identical
to the average marginal, thus implying that $\Delta_0 = 0$. With this
in mind, it is tempting to assert that the largest regret is obtained
at an i.i.d.\ distribution, since transitions from i.i.d to product, and
from product to arbitrary distribution, only subtract from the regret
value. While appealing, this is unfortunately not the case: in many
instances the final term, $\Delta_2$, can be made larger with
a non-i.i.d.\ (and even non-product) distribution, even at the added
cost of positive $\Delta_0$ and $\Delta_1$ terms,
so that $\Reg_T^\text{i.i.d.} = o(\Reg_T)$ as a function of $T$.
In some cases, however, we show that a lower bound on the regret
can be obtained with an i.i.d.\ distribution at a cost of
only a constant factor.


\section{Properties of $\Phi$}

In statistical learning, the rate of decay of prediction
error is known to depend on the curvature of the loss: more
curvature leads to faster rates (see, for
example,~\cite{LeeBarWil98importance,Men03fewnslt,BarJorMc05}),
and slow (e.g. $\Omega(T^{-1/2})$) rates occur when the
loss is not strictly convex, or when the
minimizer of the expected loss is not
unique~\cite{LeeBarWil98importance,Mendelson08lower}.  There
is a striking parallel with the behavior of the regret in
online convex optimization; again the curvature of the loss
plays a central role.  Roughly speaking, if $\ell$ is
strongly convex or exp-concave, second-order
gradient-descent methods ensure that the regret grows no
faster than $\log T$ (e.g. \cite{Hazan06}); if $\ell$ is
linear, the regret can grow no faster than $\sqrt{T}$ (e.g.
\cite{Zinkevich03}); intermediate rates can be achieved as
well if the curvature varies \cite{BarHazRak07}.

The previous section expresses regret as a sum of divergences under
$\Phi$, and that suggests that the curvature of $\Phi$ should
be an important factor in determining the rates of regret. We shall
see that this is the case: curvature of $\Phi$ leads to large regret,
while flatness of $\Phi$ implies small regret.

We will now show how properties of the loss function class determine the curvature of $\Phi$. In later sections we will show how such curvature properties lead directly to particular rates for $\Reg_T$. First, let us provide a fruitful geometric picture, rooted in convex analysis. It allows us to see the function $\Phi$, roughly speaking, as a mirror image of the function class.

\subsection{Geometric interpretation of $\Phi$}
\label{sec:geometric_interpretation}

In general, the set $\Z$ is uncountable, so care must be taken with regard to various notions we are about to introduce. We refer the reader to Chapter~10 of \cite{BorweinLewis99} for the discussion of finite vs infinite-dimensional spaces in convex analysis. Since $\Z$ is compact by assumption, we can  discretize it to a fine enough level such that the upper and lower bounds of this paper hold, as long as the results are non-asymptotic. In the present Section, for simplicity of exposition, we will suppose that the set $\Z$ is finite with cardinality $d$. This assumption is required only for the geometric interpretation; our proofs are correct as long as $\Z$ is compact.  

Hence, distributions over the set $\Z$ are associated with $d$-dimensional vectors. Furthermore, each $f\in \F$ is specified by its $d$ values on the points. We write $\ell_f\in \R^d$ for the loss vector of $f$, $\ell(\cdot,f)$. Let us denote the set of all such vectors by $\ell(\F)$. We then have

\begin{align*}  
	-\Phi(p) &= -\inf_{f \in \F} \E_p \ell (Z, f) 
	= \sup_{\ell_f \in \ell(\F)} \la -\ell_f, p \ra = \sigma_{-\ell(\F)} (p),
\end{align*}
where $\sigma_{S}(x) = \sup_{s\in S} \la s, x \ra$ is the support
function for the set $S$. 
This function is one of the most basic objects of convex analysis
(see, for instance, \cite{FundConvAnalysis}).
It is well-known that
$ \sigma_S = \sigma_{\co S} \ $; 
in other words, the support function does not change with respect to taking convex hull (see Proposition 2.2.1, page 137, \cite{FundConvAnalysis}). To this end, let us denote $S = \co [-\ell(\F)]\subset \R^d $.

\begin{figure}[h]
\centering
\includegraphics[height=2in]{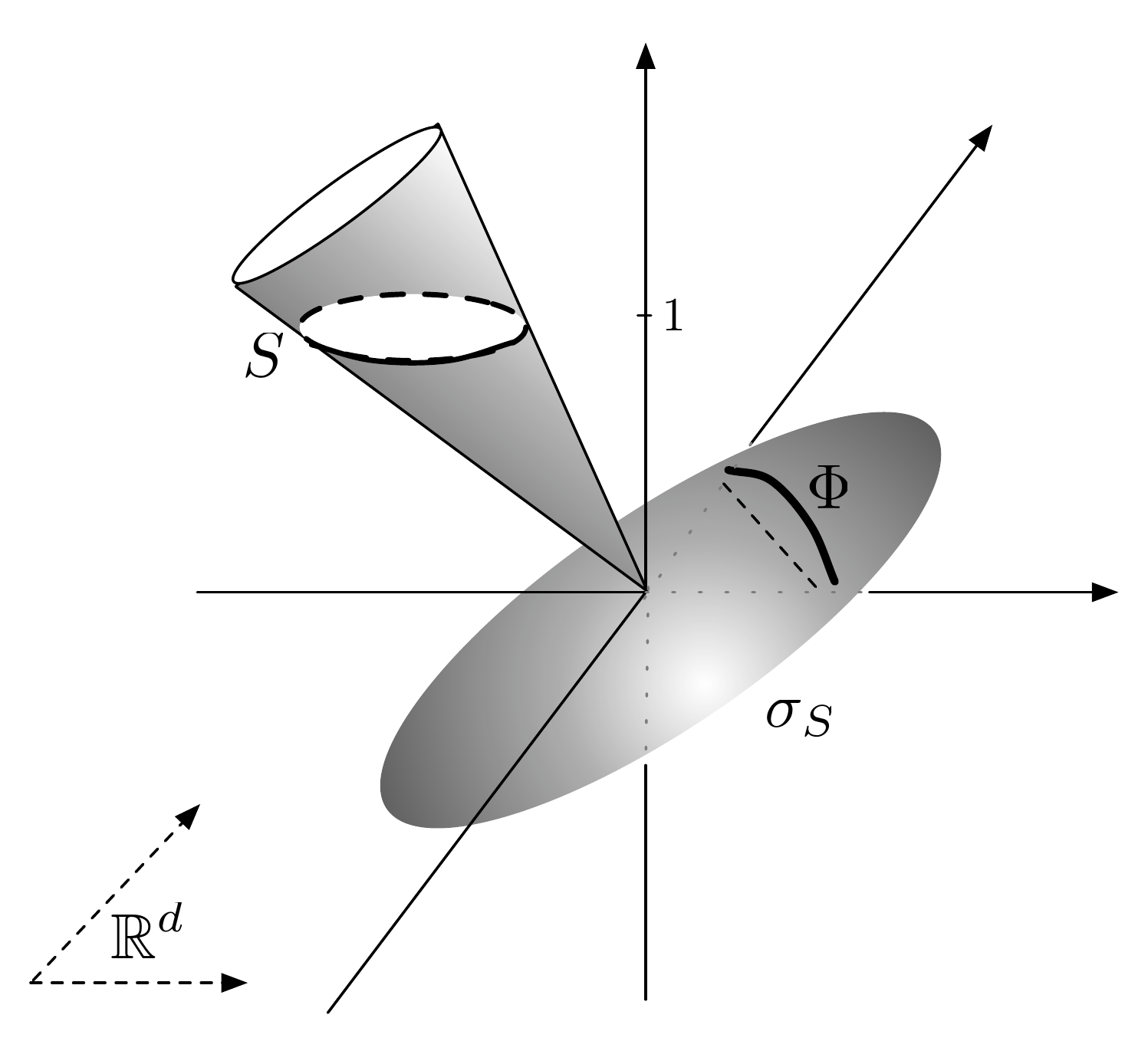}
\caption{Dual cone as the epigraph of the support function. $\Phi$ is the restriction to the simplex. }
\label{fig:duality0-3d}	
\end{figure}

It is known that the support function is {\em sublinear} and its epigraph is a cone. To visualize the support function, consider the $\R^d\times \R$ space. Embed the set $S \subset \R^d$ in $\R^d\times \{1\}$ (see Figure \ref{fig:duality0-3d}). Then construct the conic hull of $S\times \{1\}$. It turns out that the cone which is dual to the constructed conic hull is the epigraph of the support function $\sigma_S$. The dual cone is the set of vectors which form obtuse or right angles with all the vectors in the original cone. Hence, one can visualize the surface $\sigma_S$ as being at the right angles to the conic hull of $S\times \{1\}$. Now, the function $\Phi$ is just the restriction of $\sigma_S$ to the simplex. We can now deduce properties of $\Phi$ from properties of the loss class.

\subsection{Differentiability of $\Phi$}
\label{sec:phidiff}

\begin{lemma}
	The subdifferential set of $\Phi$ is the set of expected minimizers:
	$$\partial\Phi(p) = \{\ell_f:f \in \arg\min_{f\in \mathcal{F}}\E_p\ell(Z,f)\}. $$
  Hence, the functional $\Phi$ is differentiable at a distribution $p$ iff $\arg\min_{f \in \mathcal{F}}\E_p\ell(z,f)$ is unique. 
\end{lemma}
\begin{proof} 
  We have seen that $-\Phi$ is the support
  function of $\co[-\ell(\F)]$ restricted to the
  probability simplex. 
  The subdifferential set of the support function is the {\em support
  set}, that is, the set of points achieving the supremum in the
  definition of support function. 

  By examining Figure~\ref{fig:duality0-3d}, one can see why this statement is correct: roughly speaking, a point on the boundary of $S$, which supports some distribution $p$, serves as a normal to a hyperplane tangent to $\Phi$ at $p$. The precise proof of this fact, found in Proposition 2.1.5 in \cite{FundConvAnalysis}, can be extended to the infinite-dimensional case as well.

	For $\Phi$ to be differentiable, the subdifferential set has to be singleton. This immediately gives us the criterion for the differentiability of $\Phi$ stated above.
\end{proof}

In particular, for $\Phi$ to be differentiable for all distributions, the loss function class should not have a ``face'' exposed to the origin. This geometrical picture and its implications will be studied further in Section~\ref{sec:general_lower}. 

It is easy to verify that {\em strict} convexity of $\ell(z, f)$ in $f$ implies uniqueness of the minimizer for any $p$ and, hence, differentiability of $\Phi$.

\subsection{Flatness of $\Phi$ through curvature of $\ell$}
\label{sec:flatness}

In this section we show that curvature in the loss function leads to
flatness of $\Phi$. We would indeed expect such a result to hold since
regret decaying faster than $O(T^{-1/2})$ is known to occur in the
case of curved losses (e.g. \cite{BarHazRak07}), and decomposition
\eqref{eq:decomposition} suggests that this should imply flatness of
$\Phi$. More precisely, we show that if $\ell(f,z)$ is strongly convex
in $f$ with respect to some norm $\|\cdot\|$, then $\Phi$ is strongly
flat with respect to the $\ell_1$ norm on the space of distributions. Before stating the main result, we provide several definitions.

\begin{definition}
	A convex function $F$ is \emph{$\alpha$-flat} (or \emph{$\alpha$-smooth}) with respect to a norm $\| \cdot \|$ when
	\begin{equation}
	  F(y) - F(x) \leq \la \nabla F (x), y-x \ra + \alpha \|x-y\|^2
	  \label{eqn:flatdef}
	\end{equation}
	for all $x,y$. We will say that a concave function $G$ is $\alpha$-flat if $-G$ satisfies~\eqref{eqn:flatdef}. 
\end{definition}

Let us also recall the definition of $\ell_1$ (or variational) norm on distributions.
\begin{definition}
  For two distributions $p,q$ on $\Z$, we define
  $$\|p-q\|_1 = \int_{\Z} |dp(z) - dq(z)|.$$
\end{definition}
\begin{theorem}
	\label{thm:flatness}
	Suppose $\ell(z, f)$ is $\sigma$-strongly convex in $f$, that is,
	$$
	\ell\left(z, \frac{f + g}{2}\right) \leq \frac{\ell(z, f) + \ell(z,g)}{2} - \frac{\sigma}{8} \|f-g\|^2
	$$
	for any $z\in Z$ and $f,g\in \F$. Suppose further that $\ell$ is $L-Lipschitz$, that is,
	$$|\ell(z, f)-\ell(z, g)| \leq L \|f-g\|. $$
	Under these conditions, the $\Phi$-functional is $\frac{2L^2}{\sigma}$-flat  with respect to $\| \cdot \|_1$.
\end{theorem}	
The proof uses the following lemma, which shows stability of the minimizers. Its proof appears in the Appendix.
\begin{lemma}
\label{lem:stability}
	Fix two distributions $p,q$. Let $f_p$ and $f_q$ be the functions achieving the minimum in $\Phi(p)$ and $\Phi(q)$, respectively. Under the conditions of Theorem~\ref{thm:flatness},
$$\|f_p-f_q\| \leq \frac{2L}{\sigma} \|p-q\|_1.$$
\end{lemma}

\begin{proof}[Proof of Theorem~\ref{thm:flatness}]
	We have
		\begin{align}  
		  \label{eq:second_order_decomposition}
		  \Phi(p)-\Phi(q) &= \E_p \ell(z, f_p) - \E_q \ell (z, f_q)  = \left(\E_p \ell(z, f_p) - \E_q \ell (z, f_p)\right) +  \left(\E_q \ell(z, f_p) - \E_q \ell (z, f_q) \right). 
		\end{align}

		\begin{figure}[tbp]  
			\centering
				\includegraphics[width=1.8in]{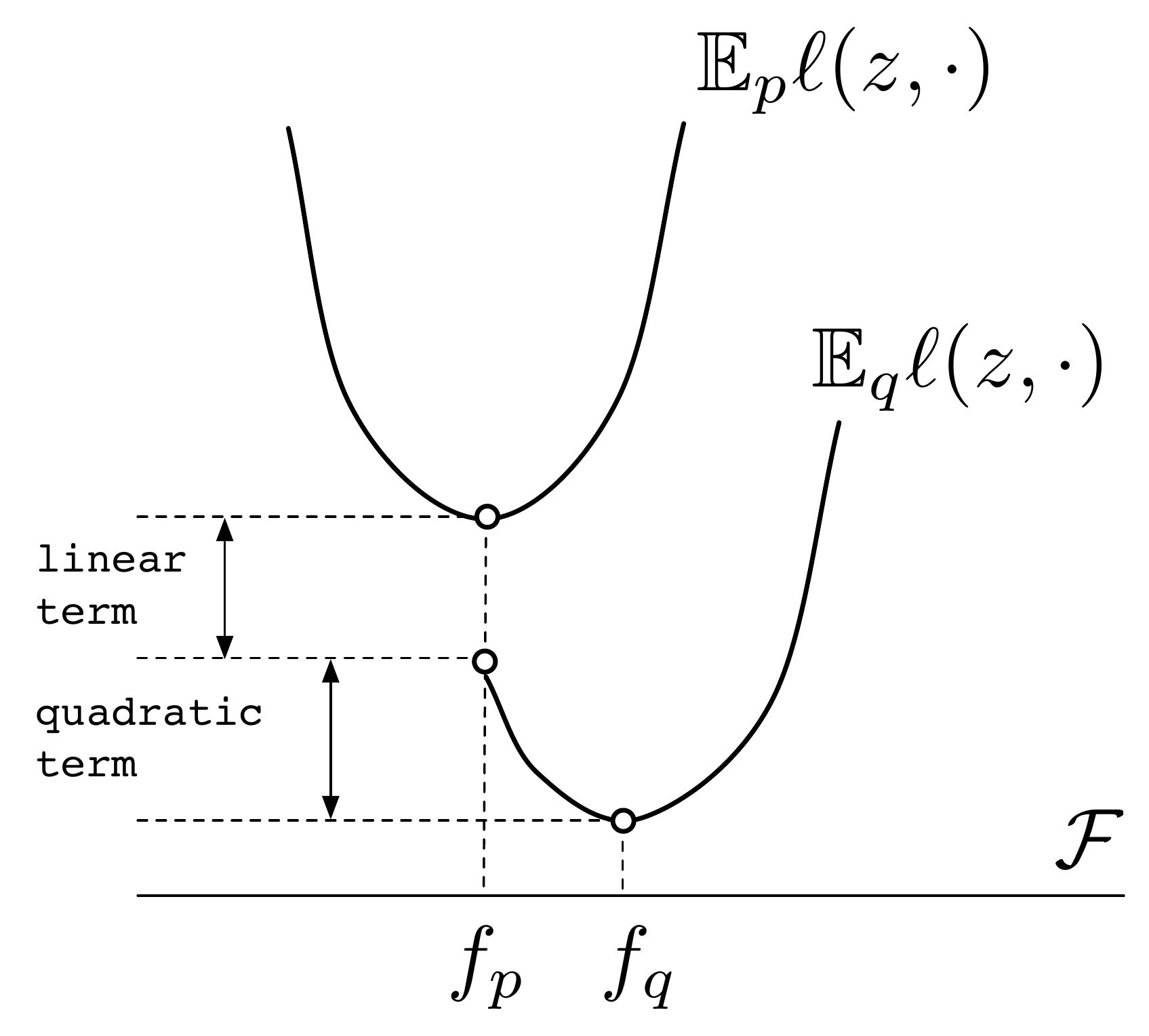}
			\caption{The two terms in the decomposition \eqref{eq:second_order_decomposition}. }
			\label{fig:decomposition}
		\end{figure}
	
	Let us first study the second term in the expression above. 
	As $f_p$ is the minimizer of $\E_p\ell(z,f)$, we have:
	$$\E_p\left[\ell(z,f_p) - \ell(z,f_q)\right] \leq 0$$
	So
	\begin{align*} 
	  \E_q\left[\ell(z,f_p) - \ell(z,f_q)\right] &\leq \E_q\left[\ell(z,f_p) - \ell(z,f_q)\right] - \E_p\left[\ell(z,f_p) - \ell(z,f_q)\right]\\
	  	&= \int(\ell(z,f_p) - \ell(z,f_q))(dq(z) - dp(z)) \\
		&\leq L\int\|f_p - f_q\||dp(z) - dq(z)| .
	\end{align*}
	Using Lemma~\ref{lem:stability}, we get:
	\begin{equation}
	  \E_q\left[\ell(z,f_p) - \ell(z,f_q)\right] \leq \frac{2L^2}{\sigma}\|p-q\|_1^2.
	  \label{eq:curvature}
	\end{equation}
	As for the first term in \eqref{eq:second_order_decomposition},
	\begin{align}
	  \E_p \ell(z, f_p) - \E_q \ell (z, f_p) &= \int_z \ell(z, f_p) (dp(z)-dq(z)) = \la \ell(\cdot, f_p), (p-q)\ra.
	\end{align}
	The fact that $\ell(\cdot, f_p)$ is a subdifferential of $\Phi$ at $p$ is proved in the appendix.
	We conclude that the first and the second terms in \eqref{eq:second_order_decomposition} are the first and the second order terms in the expansion of $\Phi$.
\end{proof}

We remark that we can arrive at above results by explicitly considering the dual function $\Phi^*$, proving strong convexity of $\Phi^*$ with respect to $\|\cdot \|_\infty$ (which follows from our assumption on $\ell$), and then concluding strong flatness of $\Phi$ with respect to $\|\cdot\|_1$. This is indeed the main intuition at the heart of our proof.


\section{Upper Bounds on $\Reg_T$}

In this section, we exhibit two general upper bounds on $\Reg_T$ that
hold for a wide class of OCO games. The first bound, which holds when
the functional $\Phi$ is differentiable and not too curved,
is of the form $\Reg_T = O(\log T)$. The second, which holds for
\emph{arbitrary} $\Phi$, e.g. where the functional may even have a
non-differentiability, is stated in terms of the \emph{Rademacher
complexity} of the class $\F$. Such Rademacher complexity results
imply a regret upper bound on the order of $\sqrt{T}$.

An intriguing observation is that these bounds are proved without
actually exhibiting a strategy for the Player, as is typically done.
This illustrates the power of the minimax duality approach:
we can prove the existence of an optimal algorithm, and determine
its performance, all without providing its construction.

Throughout, we shall refer to $O(\sqrt{T})$ rates as ``slow rates'' and $O(\log T)$ as fast rates. These notions are borrowed from the statistical learning literature, where fast rates of convergence of the empirical minimizer to the best in class arise from certain assumptions, such as convexity of the class and square loss. The slow rates, on the other hand, are exhibited by the situations where the expected minimizer of the loss is non-unique.

\subsection{Fast Rates: Exploiting the Curvature}
For differentiable $\Phi$ with bounded second derivative, we can prove
that the regret grows no faster than logarithmically in $T$. Of
course, rates of $\log T$ have been given previously \cite{Hazan06,
TakWar00, Vovk98}. We build upon these results in the present work by
showing that logarithmic regret must always arise when $\Phi$
satisfies a flatness condition.

\begin{theorem}
	\label{thm:logt_flatness}
	Suppose the $\Phi$ functional is differentiable and $\alpha$-flat with respect the norm $\| \cdot \|_1$ on $\Probs$. Then
	$\Reg_T \leq 4\alpha \log T$.
\end{theorem}

We immediately obtain the following corollary.
\begin{corollary} \label{cor:logTregret}
Suppose functions $\ell(z, f)$ are $\sigma$-strongly convex and $L-Lipschitz$ in $f$. Then
$\Reg_T \leq \frac{8L^2}{\sigma}\log T.$
\end{corollary}

Furthermore, as we show in Section~\ref{sec:quadratic}, the $\log T$
bound is tight for quadratic functions; there is an explicit joint
distribution for the adversary which attains this value.

The proof of Theorem~\ref{thm:logt_flatness} involves the following
lemma.
\begin{lemma}
	\label{lem:upper_bound_recursive}
	The $\jp$-regret can be upper-bounded as
	\begin{align*}
		\Reg_T(\jp) \leq \E \left[ \sum_{t=1}^T t\cdot
		\Dvg\left(\Unif[t],\Ub[t]\right) \right]	
	\end{align*}
	where $\Ub[t](\cdot) = \left(\frac{t-1}{t}\right)\Unif[t-1](\cdot) + \frac{1}{t}\pt$.
\end{lemma}
 
\begin{proof}
	Consider the following difference:
	\begin{align*}
		\delta_T :&= \Tinv \E\Phi\left(p_T(\cdot|Z_1^{T-1})\right) - \E\Phi\left(\Unif\right) \\
		&= \Tinv \E\Phi\left(p_T(\cdot|Z_1^{T-1})\right) -\E\Phi\left(\Ub\right)
		+ \E\Phi\left(\Ub\right) - \E\Phi\left(\Unif\right)\\
	\end{align*}
	For the first difference we use concavity of $\Phi$. The
	second difference can be written as a divergence because the linear term vanishes in expectation.
	Indeed,
 		$$\E\left\la \nabla\Phi\left( \Ub\right), \Tinv (\one_{Z_T}(\cdot) - p_T(\cdot|Z_1^{T-1}) ) \right\ra = 0$$
because the gradient does not depend on $Z_T$, while 
$$\E_{Z_T} \left[\one_{Z_T}(\cdot) | Z_1^{T-1}\right] = p_T(\cdot|Z_1^{T-1}).$$ Hence, 
	\begin{align*}
		\delta_T &\leq -\left(\frac{T-1}{T}\right)\E\Phi(\Unif[T-1]) 
		+ \E\Dvg\left(\Unif, \Ub\right)
	\end{align*}
and so
	\begin{align*}
		\Reg_T(\jp) &= \sum_{t=1}^T \E\Phi\left(p_t(\cdot|Z_1^{t-1})\right) - T\E\Phi\left(\Unif\right)\\
		&= \sum_{t=1}^{T-1} \E\Phi\left(p_t(\cdot|Z_1^{t-1})\right) + T\delta_T \\
		&\leq \sum_{t=1}^{T-1} \E\Phi\left(p_t(\cdot|Z_1^{t-1})\right) - (T-1)\E\Phi(\Unif[T-1]) 
		+ T\E\Dvg\left(\Unif, \Ub\right). 
	\end{align*}
\end{proof}

Before proceeding, note that we may interpret $\Ub[t]$ as the
conditional expectation of the uniform distribution $\Unif[t]$ given $Z_1,
\ldots, Z_{t-1}$. The flatness of $\Phi$ will allow us to show that
$\Ub[t]$ deviates very slightly from $\Unif[t]$ in
expectation---indeed, by no more than $O(\frac{1}{t^2})$. This is
crucial for obtaining fast rates: for general $\Phi$ (which may be
non-differentiable), it is natural to expect
$\Dvg\left(\Unif[t],\Ub[t]\right) = \Omega(1/t)$. In this case,
the regret would be bounded by $O(\sum_t t \cdot 1/t) = O(T)$,
rendering the above lemma useless.
 
\begin{proof}[{\bf Proof of Theorem \ref{thm:logt_flatness}}]
	We have that the divergence terms in Lemma~\ref{lem:upper_bound_recursive} are bounded as
	\begin{align*} 
	  t\cdot\Dvg&\left(\Unif[t], \Ub[t]\right) 
	  \leq t\alpha \left\|\frac{1}{t}\one_{Z_t}(\cdot) - \frac{1}{t}p_t(\cdot|Z_1^{t-1})\right\|^2_1 \leq \frac{4\alpha}{t}
	\end{align*}
	because the variational distance between distributions is bounded by $4$:
	$$\left(\int_z \left|\delta_{Z_t}(z) - dp_t(z|Z_1^{T-1})\right|\right)^2 \leq 4.$$
	
\end{proof}

\subsection{General $\sqrt{T}$ Upper Bounds}

We start with the definition of Rademacher averages, one of the central notions of complexity of a function class.
\begin{definition}
	Denote by 
	  $$
	  \Rad_T(\ell(F)) := \frac{1}{\sqrt{T}} \E_{\epsilon_1^T}
	  \left(\sup_{f\in\F} \left|\sum_{t=1}^T \epsilon_t\ell(f, Z_t)
	  \right|\right)
	  $$
	the {\em data-dependent Rademacher averages}
	of the class $\ell(\F)$. Here, $\epsilon_1\ldots \epsilon_T$ are
	independent Rademacher random variables (uniform on
	$\{\pm 1\}$).
\end{definition}
We will omit the subscript $T$ and dependence
on $Z_1^T$, for the sake of simplicity.
In statistical learning theory, Rademacher averages often provide the
tightest guarantees on the performance of empirical risk
minimization and other methods. The next result
shows that the Rademacher averages play a key role in online convex
optimization as well, as the minimax regret is upper bounded by the
worst-case (over the sample) Rademacher averages. In the next section,
we will also show lower bounds in terms of Rademacher averages for
certain linear games, showing that this notion of complexity is
fundamental for OCO.

\begin{theorem}
	\label{thm:rademacher_upper}
	$$ \Reg_T \leq 2\sqrt{T}\sup_{Z_1^T\in \Z^T} \Rad (\ell(\F)).$$
\end{theorem}
\begin{proof}
Let $\jp$ be an arbitrary joint distribution. Let $\hat{f}$ be an empirical minimizer over $Z_1^T$, a sequence-dependent function.
Then 
\begin{align*}  
\Tinv\Reg_T (\jp) = \E \Tinv \sum_{t=1}^T \left[\Phi(\pt)- \Phi(\Unif) \right]
\leq \E \Tinv \sum_{t=1}^T \left[ \E_{\pt} \ell(Z, \hat{f}) - \Tinv\sum_{s=1}^T \ell(Z_s, \hat{f})  \right],
\end{align*}
as the particular choice of $\hat{f}$ is (sub)optimal. 
Replacing the $\hat{f}$ by the supremum over $\F$,
\begin{align*}
\Tinv\Reg_T (\jp) &\leq \E \Tinv \sum_{t=1}^T \left[ \E_{\pt} \ell(Z, \hat{f}) -  \ell(Z_t, \hat{f})  \right]\\
&\leq \E \sup_{f\in \F}\Tinv \sum_{t=1}^T \left[ \E_{\pt} \ell(Z, f) -  \ell(Z_t, f)  \right] \\
&= \E \sup_{f\in \F}\Tinv \sum_{t=1}^T \left[ \E_{\pt} \ell(Z'_t, f) -  \ell(Z_t, f)  \right] \\
&\leq \E \sup_{f\in \F}\Tinv \sum_{t=1}^T \left[ \ell(Z'_t, f) -  \ell(Z_t, f)  \right],
\end{align*}  
where we renamed each dummy variable $Z$ as $Z'_t$.
Even though $Z_t$ and $Z'_t$ have the same conditional expectation, we
cannot generally exchange them keeping the distribution of the whole
quantity intact. Indeed, the conditional distributions for $\tau>t$
will depend on $Z_t$ and not on $Z'_t$. The trick is to exchange them
one by one\footnote{We thank Ambuj Tewari for pointing out a mistake
in our original proof. We refer to \cite{KarTew09unpub} for a similar analysis.}, starting from $t=T$ and going backwards,
introducing an additional supremum. (One can view the sequence $\{Z'_t\}$ as being {\em tangent} to $\{Z_t\}$ (see \cite{delaPenaGine98}).)
To this end, for any fixed $\epsilon_T\in \{-1,+1\}$,
\begin{align*}  
	&\E \sup_{f\in \F}\Tinv \sum_{t=1}^T \left[ \ell(Z'_t, f) -  \ell(Z_t, f)  \right]\\ 
	&= \E \sup_{f\in \F}\left(\Tinv \sum_{t=1}^{T-1} \left[ \ell(Z'_t, f) -  \ell(Z_t, f) \right]  
	+\Tinv\epsilon_T\left(\ell(Z'_T, f) -  \ell(Z_T, f)\right)\right)
\end{align*}
because for the last step, indeed, $Z_T$ and $Z'_T$ can be exchanged. Since this holds for any $\epsilon_T$, we can take it to be a Rademacher random variable. Thus,
\begin{align*}
	&\E \sup_{f\in \F}\left(\Tinv \sum_{t=1}^{T-1} \left[ \ell(Z'_t, f) -  \ell(Z_t, f) \right] 
	+ \Tinv\epsilon_T\left(\ell(Z'_T, f) -  \ell(Z_T, f)\right)\right)\\
	&\leq \E_{\epsilon_T}\E \sup_{f\in \F}\left(\Tinv \sum_{t=1}^{T-1} \left[ \ell(Z'_t, f) -  \ell(Z_t, f) \right] 
	+ \Tinv\epsilon_T\left(\ell(Z'_T, f) -  \ell(Z_T, f)\right)\right)\\
	&\leq \sup_{Z_T, Z'_T} \E_{Z_1^{T-1}}\E_{\epsilon_T} \sup_{f\in \F}\left(\Tinv \sum_{t=1}^{T-1} \left[ \ell(Z'_t, f) -  \ell(Z_t, f) \right]
	+\Tinv\epsilon_T\left(\ell(Z'_T, f) -  \ell(Z_T, f)\right)\right),
\end{align*}
where we assumed the worst case over $Z_T,Z'_T$. The first expectation is now taken over the shorter sequence $1,\ldots,T-1$. 
Repeating the process, we have that $\Tinv\Reg_T(\jp)$ is bounded by
\begin{align*}
	\Tinv\Reg_T(\jp)  &\leq \sup_{Z_1^T, Z_1^{'T}} \E_{\epsilon_1^T}
	  \sup_{f\in \F}\left(\Tinv \sum_{t=1}^T
	  \epsilon_t\left(\ell(Z'_t, f) -  \ell(Z_t, f)\right)\right)
	  \\
	  & \leq 2\sup_{Z_1^T} \E_{\epsilon_1^T} \sup_{f\in \F}
	  \Tinv \left| \sum_{t=1}^T  \epsilon_t \ell(Z_t, f)\right| = 2\frac{1}{\sqrt{T}} \sup_{Z_1^T} \Rad(\ell(F)).
\end{align*}
\end{proof}

Properties of Rademacher averages are well-known. For instance, the Rademacher averages of a function class coincide with those of its convex hull. Furthermore, if $\ell$ is Lipschitz, the complexity of $\ell(\F)$ can be upper bounded by the complexity of $\F$, multiplied by the Lipschitz constant. For example, we can immediately conclude that if the loss function is Lipschitz and the function class is a convex hull of a finite number $M$ of functions, the minimax value of the game is bounded by $ \Reg_T \leq C\sqrt{T\log M}$ for some constant $C$. Similarly, a class with VC-dimension $d$ would have $\log M$ replaced by $d$. Theorem~\ref{thm:rademacher_upper} is, therefore, giving us the flexibility to upper bound the minimax value of OCO for very general classes of functions.

Finally, we remark that most known upper bounds on Rademacher averages
do not depend on the underlying distribution, as they hold for the
worst-case empirical measure (see \cite{Men03fewnslt}, p. 27). Thus,
the supremum over the sequences might not be a hinderance to using
known bounds for $\Rad(\ell(\F))$.

\subsection{Linear Losses: Primal-Dual Ball Game}
\label{sec:primal_dual_upper}

Let us examine the linear loss more closely. Of particular interest are linear games when $\F=B_{\|\cdot\|_*}$ is a ball in some norm $\|\cdot\|_*$ and $\Z=B_{\|\cdot\|}$, the two norms being dual.  For this case, Theorem~\ref{thm:rademacher_upper} gives an upper bound of 
\begin{align}
	\label{eq:upper_bound_for_linear_sup_Z}
  \Tinv \Reg_T &\leq 2\sup_{Z_1^T} \E_{\epsilon_1^T}\sup_{f\in\F} f^\tr\left(\Tinv\sum_{t=1}^T \epsilon_t Z_t\right) 
  = 2\sup_{Z_1^T} \E_{\epsilon_1^T} \left\|\Tinv\sum_{t=1}^T \epsilon_t Z_t \right\|.
\end{align}
Fix $Z_1\ldots Z_{T-1}$ and observe that the expected norm is a convex function of $Z_T$. Hence, the supremum over $Z_T$ is achieved at the boundary of $\Z$. The same statement holds for all $Z_t$'s. Let $z_1^*,\ldots,z_T^*$ be the sequence achieving the supremum. Now take a distribution for round $t$ to be $p^*_t(z)= \frac{1}{2}\left(\one_{z_t^*}(\cdot)+\one_{-z_t^*}(\cdot)\right)$ and let $\jp^* = p_1^*\times\ldots\times p_T^*$ be the product distribution. It is easy to see that 
\begin{align} 
	\label{eq:upper_bound_for_linear_no_sup}
	\Tinv \Reg_T &\leq 2\sup_{Z_1^T} \E_{\epsilon_1^T} \left\|\Tinv\sum_{t=1}^T \epsilon_t Z_t \right\| = 2\E_{\epsilon_1^T} \left\|\Tinv\sum_{t=1}^T \epsilon_t z^*_t \right\| \nonumber\\
	&= 2\E_{\jp^*} \E_{\epsilon_1^T} \left\|\Tinv\sum_{t=1}^T \epsilon_t Z_t \right\| = \frac{2}{\sqrt{T}}\E_{\jp^*} \Rad(\F).
\end{align}
Also note that $p^*_t$ has zero mean. It will be shown in Section~\ref{sec:linear_lower} that the lower bound arising from this distribution is 
$$\E_{\jp^*} \E_{\epsilon_1^T} \left\|\Tinv\sum_{t=1}^T \epsilon_t Z_t \right\| = \frac{1}{\sqrt{T}}\E_{\jp^*} \Rad(\F),$$
which is only a factor of $2$ away.
Thus, the adversary can play a product distribution that arises from the maximization in \eqref{eq:upper_bound_for_linear_sup_Z} and achieve regret at most a factor 2 from the optimum.



\section{$\Omega(\sqrt{T})$ bounds for non-differentiable $\Phi$}
\label{sec:general_lower}

In this section, we develop lower bounds on the minimax value $\Reg_T$ based on the geometric view-point described in Section~\ref{sec:geometric_interpretation}. 

Theorem~\ref{thm:logt_flatness} shows that the regret is upper bounded
by $\log T$ for the case of strongly convex losses, and this upper
bound is tight if the loss functions are quadratic, as we show later
in the paper. Thus, flatness of $\Phi$ implies low regret. What about
the converse? It turns out that if $\Phi$ is non-differentiable (has a
point of infinite curvature), the regret is lower-bounded by
$\sqrt{T}$, and this rate is achieved with $\jp =
p^T$, where $p$ corresponds to a point of non-differentiability of
$\Phi$.

The geometric viewpoint is fruitful here: vertices (points of
non-differentiability) of $\Phi$ correspond to \emph{exposed faces} in
the loss class $S = \co [-\ell(\F)]$ (see Figure~\ref{fig:duality0-3d})
suggesting that the lower bounds of
$\Omega(\sqrt{T})$ arise from having two distinct minimizers of
expected error---a striking parallel to the analogous results for
stochastic settings~\cite{LeeBarWil98importance,Mendelson08lower}.

\begin{figure}[h] 
		\centering
		\includegraphics[height=1.4in]{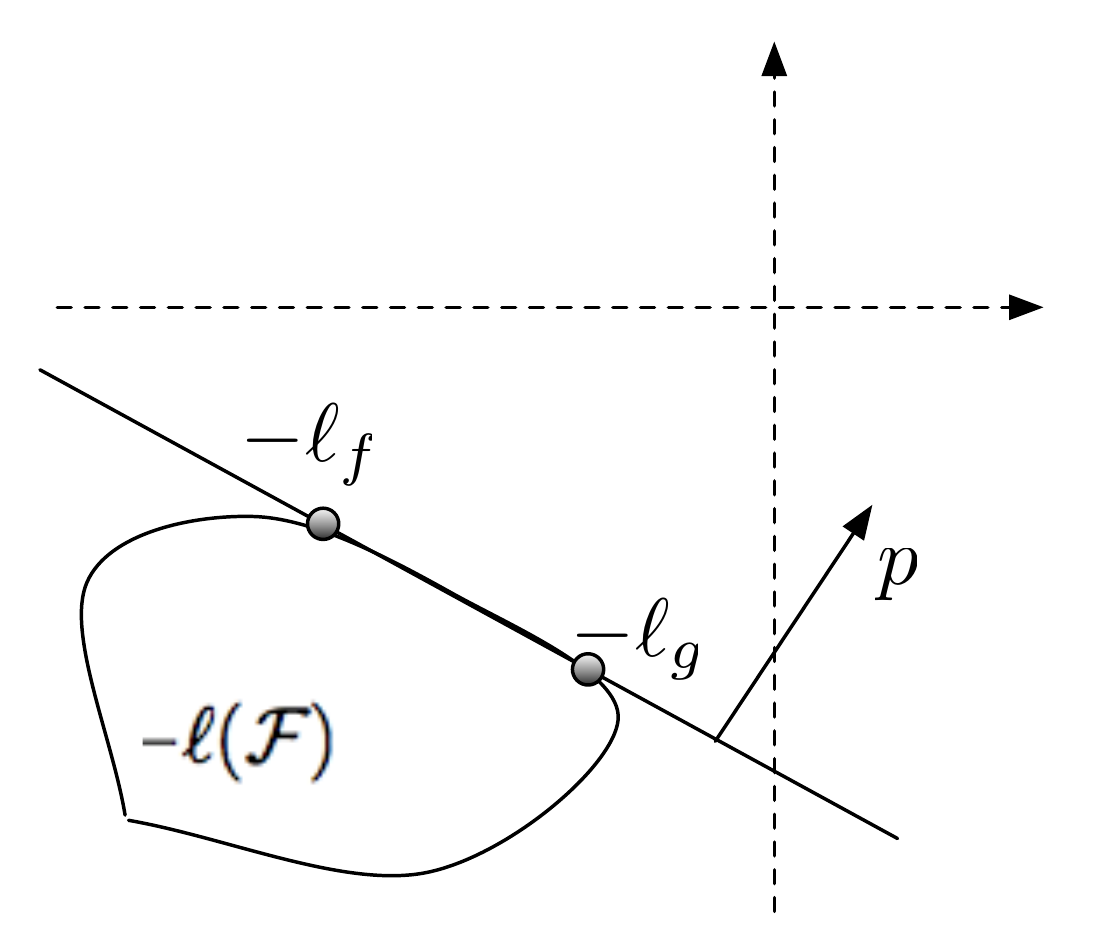}
		\caption{The face of the convex hull of the loss class is supported by a probability distribution $p$.}
		\label{fig:support}	
\end{figure}

To be more precise, vertices of $\sigma_S$ (and $\Phi$) translate into flat parts (non-singleton {\it exposed faces}) of $S (\co [-\ell(\F)])$  and the other way around. Corresponding to an exposed face is a supporting hyperplane. If $\ell(\F)$ is non-negative, then any exposed face facing the origin is supported by a hyperplane with positive co-ordinates (which can be normalized to get a distribution). So a non-singleton face exposed to the origin is equivalent to having at least two distinct minimizers $f$ and $g$ of $\E_p \ell(Z,\cdot)$ for some $p$,  as discussed in Section~\ref{sec:phidiff}. We demonstrate this scenario, along with the supporting hyperplane, $\la \ell_f, p \ra = \la \ell_g, p \ra$, in Figure \ref{fig:support}.

Suppose there is a non-singleton face $F_S(p)$ of $\co [-\ell(\F)]$ supported by some distribution $p$. It is known (see \cite{FundConvAnalysis}) that any exposed face is a convex set. The extreme points of this convex set are vectors $\ell_f$ (and not convex combinations such as $\frac{\ell_f + \ell_g}{2}$). Furthermore, for all $\ell_f, \ell_g \in F_S(p)$, we have  $\la \ell_f, p \ra = \la \ell_g, p \ra$. 

Define the set of expected minimizers
under $p$ as
  \[
  \F^*:=\{f\in\F: \E_p\ell(Z, f) = \inf_{f\in\F} \E_p\ell(Z, f) \}.
  \]
Thus, $-\ell(\F^*)\subseteq F_S(p) \subseteq \co [-\ell(\F)]$.
The lower bound we are about to
state arises from fluctuations of the empirical process over the set
$\F^*$. To ease the presentation, we will refer to the sample average
$\Tinv\sum_{t=1}^T\ell(Z_t,f)$ as $\hat\E\ell(Z,f)$.

\begin{theorem}
	\label{thm:gaussian_lower}
	Suppose $F_S(p)$ is a non-singleton face of $\co[-\ell(\F)]$, supported by $p$ (i.e. $|\F^*|>1$). Fix any
	$f^*\in\F^*$ and let $Q\subseteq \ell(\F^*)$ be any subset
	containing $\ell(\cdot, f^*)$.
	Define $\bar{Q}=\{g -\ell(\cdot, f^*): g\in
	Q\}$, the shifted loss class. Then for $T > T_0(\F)$, 
	\begin{align*}
		\Tinv\Reg_T & \geq \Tinv \Reg_T(p^T) \\
		& = \E \sup_{f\in\F^*} \left[ \E_p \ell(Z, f) - \hat{\E}\ell(Z, f) \right]\\
		& \geq \frac{c}{\sqrt{T}}\sup_{Q\subseteq \ell(\F^*)} \E \sup_{q\in \bar{Q}} G_q,
	\end{align*}
	where $G_q$ is the Gaussian process indexed by the (centered)
	functions in $\bar Q$, and $c$ is some absolute constant.
\end{theorem}
\begin{proof}
Recalling that $\E\ell(Z, f) = \inf_{g\in\F} \E\ell(Z, g) = \Phi(p)$ for all $f\in\F^*$, we have
\begin{align*}
	\Tinv\Reg_T &\geq \Tinv \Reg_T(p^T) =  \Phi(p)-\E_p \Phi(\Unif) \\
	&= \Phi(p) - \E \inf_{f\in\F} \Tinv\sum_{t=1}^T \ell(Z_t, f)\\
	&\geq \Phi(p) - \E \inf_{f\in\F^*} \hat{\E}\ell(Z, f)\\
	&= \E \sup_{f\in\F^*} \left[ \E_p \ell(Z, f) - \hat\E\ell(Z, f) \right] \\
	&\geq \sup_{Q\subseteq \ell(\F^*)} \E \sup_{f: \ell_f \in Q} \left[ \E_p \ell(Z, f) - \hat\E\ell(Z, f) \right]
\end{align*}

Now, fix any $f^*\in\F^*$. The proof of Theorem 2.2 in \cite{LecMen09}
reveals that empirical fluctuations are lower bounded by the supremum
of the Gaussian process indexed by $\bar{Q}$. To be precise,
there exists $T_0(\F)$ such that for
$T>T_0(\F)$ with probability greater than $c_1$,
$$\inf_{f:\ell_f\in Q} \hat\E (\ell(Z, f)-\ell(Z, f^*)) \leq -c_2 \frac{\E\sup_{q\in \bar{Q}} G_q}{\sqrt{T}},$$
for some absolute constants $c_1,c_2$. Rearranging and using the fact
that $\E\ell(Z, f)-\E\ell(Z, f^*) = 0$ for $f\in\F^*$,
\begin{align*} 
  \sup_{f:\ell_f\in Q} &\left[ \E\ell(Z, f) - \hat\E \ell(Z, f) + \hat\E\ell(Z, f^*)- \E\ell(Z,f^*)\right]
  \geq c_2 \frac{\E\sup_{q\in \bar{Q}} G_q}{\sqrt{T}}
\end{align*}
with probability at least $c_1$. The supremum is non-negative because $f^*\in Q$ and therefore
\begin{align*} 
  \E \sup_{f:\ell_f\in Q} &\left[ \E\ell(Z, f) - \hat\E\ell(Z, f) \right]
  \geq c_1c_2 \frac{\E\sup_{q\in \bar{Q}} G_q}{\sqrt{T}}.
\end{align*}

\end{proof}

We remark that in the experts case, the lower bound on regret becomes $\sqrt{T\log N}$, as the Gaussian process reduces to $N$ independent Gaussian random variables. We discuss this and other examples in the next section.

\section{Lower Bounds for Special Cases}
\label{sec:lower}

We now provide lower bounds for particular games. Some of the
results of the section are known: we show how the proofs follow from the
general lower bounds developed in the previous section.

\subsection{Linear Loss: Primal-Dual Ball Game}
\label{sec:linear_lower}

Here, we develop lower bounds for the case considered in
Section~\ref{sec:primal_dual_upper}. As before, to prove a lower bound
it is enough to take an i.i.d.\ or product distribution. In particular, the product distribution described after Eq.~\eqref{eq:upper_bound_for_linear_sup_Z} is of particular interest. To this end, choose $\jp=p_1\times\ldots\times p_T$ to be a product of {\em symmetric} distributions on the surface of the primal ball $\Z$ with $\E_{p_t} Z = 0$. We conclude that $\Phi(p_t)=0$ and 
\begin{align*} 
  \Tinv \Reg_T \geq -\E\Phi(\Unif) &= -\E\inf_{f\in \F} f\cdot \left(\Tinv\sum_{t=1}^T Z_t \right) 
	= \E \left\|-\Tinv\sum_{t=1}^T Z_t \right\| 
\end{align*}
by the definition of dual norm. Now, because of symmetry,
\begin{align} 
	\label{eq:linear_primal_dual_norm}
	\E \left\|-\Tinv\sum_{t=1}^T Z_t \right\| = \Tinv \E_\epsilon\E \left\|\sum_{t=1}^T \epsilon_t Z_t \right\| = \Tinv\E  \E_\epsilon\left\|\sum_{t=1}^T \epsilon_t Z_t \right\|.
\end{align}

We conclude that $\Reg_T \geq \sqrt{T}\E\Rad(\F)$, the expected Rademacher averages of the dual ball acting on the primal ball. This is within a factor of $2$ of the upper bound \eqref{eq:upper_bound_for_linear_no_sup} of Section~\ref{sec:primal_dual_upper}. Hence, for the linear game on primal-dual ball, a product distribution is within a factor $2$ from the optimum. Note that this is not true for curved losses.

Now, consider the particular case of $\F=\Z=B_2$, the Euclidean ball. We will consider three distributions $\jp$.
\begin{itemize}
	\item Suppose $\jp$ is such that $\pt$ puts mass on the intersection of $B_2$ and the subspace perpendicular to $\sum_{s=1}^{t-1} Z_s$ and $\E [Z|Z_1^{t-1}] = 0$. Then 
$\E \left\|\sum_{t=1}^T Z_t \right\| = \sqrt{T}$ by unraveling the sum from the end. In fact, this is shown to be the optimal value for this problem in \cite{AbeBarRakTew08colt}. We conclude that a non-product distribution achieves the optimal regret for this problem. 
	\item Consider any symmetric i.i.d.\ distribution on the surface of the ball $\Z$. Note that for this case we still have the lower bound of Eq.~\eqref{eq:linear_primal_dual_norm}. Kinchine-Kahane inequality then implies $\Reg_T \geq \sqrt{\frac{T}{2}}$ and the constant $\sqrt{2}$ is optimal (see \cite{LatOli94}) in the absence of further assumptions.
	\item Consider another example of an i.i.d.\ distribution
	that puts equal mass on two points $\pm z_0$ on each
	round, with $\|z_0\|=1$. It then follows that this i.i.d.\ distribution achieves the regret equal to the length of the random walk $\E\left|\sum_{t=1}^T \epsilon_t \right|$, which is known to be {\em asymptotically}
	$\sqrt{2T/\pi}$.
	\item We expect that putting a uniform distribution on the surface of the ball will give a regret close to optimal $\sqrt{T}$ as the number of dimensions grows, since $Z_t$ is likely to be orthogonal to the sum of previous choices, as in the first (dependent) example.	
\end{itemize}
 We conclude that for the Euclidean game, the best strategy of the adversary is a sequence of dependent distributions, while product and i.i.d.\  distributions come within a multiplicative constant close to 1 from it.

\subsection{Experts Setting}

The experts setting provides some of the easiest examples for linear games. We start with a simplified game, where $\F=\Z=\Delta_N$, the $N$-simplex. The $\Phi$ function for this case is easy to visualize. We then present the usual game, where the set $\Z=[0,1]^N$. In both cases, we are interested in lower-bounding regret.

\subsubsection{The simplified game}
Let us look at the game when only one expert can suffer a loss of $1$ per round, i.e. the space of actions $\Z$ contains $N$ elements $e_1,\ldots, e_N$. The probability over these choices of the adversary is an $N$-dimensional simplex, just as the space of functions $\F$. For any $p\in \Delta_N$,
\begin{align*}
  \Phi (p) = \min_{f\in\Delta_N} \E_p Z\cdot f = \min_f p\cdot f = \min_{i\in [N]} p_i
\end{align*}
and therefore the $\Phi$ has the shape of a pyramid with its maximum at $p^*=\frac{1}{N}{\mathbf 1}$ and $\Phi(p^*)=1/N$. The regret is lower-bounded by an i.i.d game with this distribution $p^*$ at each round, i.e.
\begin{align*}
  \Reg_T \geq \Phi(p^*) - \E \Phi (U) &= \frac{1}{N} - \E\min_{f\in\Delta_N} \left( \Tinv\sum Z_t\right)f\\ 
 &= \E \max_{i\in [N]} \left[ \frac{1}{N}-\frac{n_i}{T}\right],
\end{align*}
where $n_i$ is the number of times $e_i$ has been chosen out of $T$ rounds.
This is the expected maximum deviation from the mean of a multinomial distribution, i.e. $1/N$ minus the smallest proportion of balls in any bin after $T$ balls have been distributed uniformly at random.

To obtain the lower bound on the maximum deviation, let us turn to Section~\ref{sec:general_lower}. The convex hull of the (negative) loss class $\co[-\ell(\F)]$ is the simplex itself. This is also the face supported by the uniform distribution $p^*$. The lower bound of Theorem~\ref{thm:gaussian_lower} involves the Gaussian process indexed by a set $Q$. Let us take $f^*=\frac{1}{N}{\mathrm 1}$ and $\F^*=\{e_1,\ldots,e_N\}\cup\{f^*\}$. We can verify that $\E e_i^\tr Z = \Phi(p^*) = \frac{1}{N}$, the covariance of the process indexed by $Q=\ell(\F^*)$ is $\E (e_i^\tr Z - \frac{1}{N})(e_j^\tr Z- \frac{1}{N}) = -\frac{1}{N^2}$ for $i\neq j$ and the variance is $\E (e_i^\tr Z-\frac{1}{N})^2 = \frac{N-1}{N^2}$. Let $\{Y_i\}_1^N$ be the Gaussian random variables with the aforementioned covariance structure. Then $\|Y_i-Y_j\|^2 = \E(Y_i-Y_j)^2 = \frac{2}{N}$. We can now construct independent Gaussian random variables $\{X_i\}_1^N$ with the same distance by putting $\frac{2}{N}$ on the diagonal of the covariance matrix. By Slepian's Lemma, $\frac{1}{2}\E \sup_{i} X_i \leq \E \sup_i Y_i$, thus giving us the lower bound 
$$ \Reg_T \geq c\sqrt{\frac{T\log N}{N}}$$
for this problem, for some absolute constant $c$ and $T$ large enough.

\subsubsection{The general case}

In the more general game, any expert can suffer a $0/1$ loss. Thus,
$p$ is a distribution on $2^N$ losses $Z$. To lower bound the regret,
choose a uniform distribution on $2^N$ binary vectors as the i.i.d.\ choice for the adversary. We have
$\Phi\left(\frac{1}{2^N}{\mathbf 1}\right) = \min_{f\in \Delta_N} f\cdot\E Z = 1/2.$
As for the other term,  $\E \Phi(\Unif) = \E \min_{f\in \Delta_N} f\cdot \left( \Tinv\sum Z_t\right).$
Thus, the regret is 
$$ \Tinv \Reg_T \geq \E \max_{i\in[N]} \left[ \frac{1}{2} - \frac{\sum_{t=1}^T \epsilon_{i,t}}{T}\right],$$
where $\epsilon_{i,t}$ are Rademacher $\{\pm 1\}$-valued random
variables. It is easy to show that the expected maximum is lower
bounded by $c\sqrt{\log N/T}$. This coincides with a result
in~\cite{CesaBianchiLugosi06book}, which shows that the asymptotic
behavior is $\sqrt{\log N/(2T)}$.


\subsection{Quadratic Loss}
\label{sec:quadratic}

We consider the quadratic loss, $\ell(z, f) = \|f-z\|^2$. This loss function is $1$-strongly convex, and therefore we already have the $O(\log T)$
bound of Corollary~\ref{cor:logTregret}. In this section, we present an \emph{almost} matching lower bound using a particular adversarial strategy. The problem of quadratic loss was previously addressed in \cite{TakWar00}; we reprove their lower bound in our framework, borrowing a number of tricks from that work.

Following Section~\ref{sec:general_lower}, it is tempting to use an
i.i.d.\ distribution and compute the regret explicitly. Unfortunately,
this only leads to a constant lower bound, whereas we would hope to
match the upper bound of $\log T$. We can show this easily: let $\jp := p^T$ be some i.i.d.\ distribution, then
\begin{align*}  
	&T\E \Phi (\Unif) = (T-1) \E \|Z_1\|^2 - \frac{T(T-1)}{T} \E\la Z_1,Z_2\ra \\
 	&= (T-1)\left( \E\|Z\|^2 - (\E Z)^2 \right) = (T-1)\mbox{var} (Z) \\
	&= (T-1)\Phi(p).
\end{align*}
Thus
$$\Reg(p^T) = T \Phi(p) - T \E \Phi (\Unif) = \Phi(p),$$ 
where we see that the last term is independent of $T$. 

Indeed, obtaining $\log T$ regret requires that we look further than i.i.d. To this end, define
\begin{align*}
	c_T & := \Tinv \\
	c_{t-1} & := c_t + c_t^2  \quad \text{ for all } t = T, T-1, \ldots, 2
\end{align*}
We construct our distribution $\jp$ using this sequence as follows. Assume $\Z = \F = [-1,1]$ and for convenience let $Z_{1:s} := \sum_{t=1}^s Z_t$. Also, for this section, we use a shorthand for the conditional expectation, $\E_t[\cdot] := \E[\cdot | Z_1, \ldots, Z_{t-1}]$. Each conditional distribution is chosen as
\[
	p_t(Z_t = z | Z_1, \ldots, Z_{t-1}) := \begin{cases}
		\frac{1 + c_t Z_{1:t-1}}{2}, &\text{ for }z = 1 \\
		\frac{1 - c_t Z_{1:t-1}}{2}, &\text{ for }z = -1 \\		
	\end{cases}.
\]
Notice that this choice ensures that $\E_t Z_t = c_t Z_{1:t-1}$, i.e. the conditional expectation is identical to the observed sample mean scaled by some  \emph{shrinkage factor} $c_t$. That $\frac{1 + c_t Z_{1:t-1}}{2} \in [0,1]$ follows from the statement $c_t \leq \frac{1}{t}$ which is proven by an easy induction. We now recall a result shown in \cite{TakWar00}:
\begin{lemma}[from \cite{TakWar00}] \label{lem:logTseries}
	\[
		\sum_{t=1}^T c_t = \log T - \log \log T + o(1)
	\]
\end{lemma}
This crucial lemma leads directly to the main result of this section.
\begin{theorem}
	With $\jp$ defined above, $\Reg_T(\jp) = \sum_{t=1}^T c_t$ and therefore
	\[
		\Reg_T(\jp) = \log T - \log \log T + o(1)
	\]
\end{theorem}
\begin{proof}
	For all $t=0, 1, \ldots , T$, let
	\begin{align*}
		Q_t &:= \E \left[ \sum_{s=1}^t  (Z_s - \E_s Z_s)^2 + c_t Z_{1:t}^2 - \sum_{s=1}^t Z_s^2 \right.\\
		  &\hspace{1in}\left.+ (c_{t+1} + c_{t+2} + \ldots + c_T )\right].
	\end{align*}
	We will show by a backwards induction that $Q_t = \Reg_T(\jp)$, from which the result will follow since $\sum_{t=1}^T c_t = Q_0$. 
	
	We begin with the base case, $Q_T = \Reg_T(\jp)$. Recall, $\min_f \E (f - Z)^2 =  \E (Z - \E Z)^2$. At the same time, $\min_f \sum_{t=1}^T (f - Z_t)^2 = \sum_t Z_t^2 - \frac{(\sum_t Z_t)^2}{T}$. This implies that
	\[
		\Reg_T(\jp) = \E \left[ \E_t (Z_t - \E_t Z)^2 + \frac{(\sum_t Z_t)^2}{T} -  \sum_t Z_t^2 \right] = Q_T,
	\]
	noting that the conditional expectations $\E_t[\cdot]$ are unnecessary within the full $\E[\cdot]$.
	
	We now show that $Q_t = Q_{t-1}$. To begin, we will need to compute the following conditional expectation:
	\begin{align*}
		\E_t \left[ Z_{1:t}^2 \right] &= \frac{1 + c_t Z_{1:t-1}}{2}(Z_{1:t-1} + 1)^2 \\
		&+ \frac{1 - c_t Z_{1:t-1}}{2}(Z_{1:t-1}-1)^2 = Z_{1:t-1}^2 (1 + 2 c_t) + 1
	\end{align*}
	Notice that we may write $Q_t - Q_{t-1}$ with the expression
	\begin{align*}
		& \E \left[ c_t Z_{1:t}^2 - c_{t-1}Z_{1:t-1}^2 + (Z_t - \E_t Z_t)^2 -  Z_t^2 - c_t \right]\\
		& =  \E \left[ \E_t \left[  c_t Z_{1:t}^2  + (Z_t - \E_t Z_t)^2 -  Z_t^2 \right] - c_{t-1}Z_{1:t-1}^2 - c_t \right]\\
		& = \E \left[  c_t(Z_{1:t-1}^2 (1 + 2 c_t) + 1)  - (\E_t Z_t)^2- c_{t-1}Z_{1:t-1}^2 - c_t \right]\\
		& = \E \left[  c_t Z_{1:t-1}^2 + 2 c_t^2 Z_{1:t-1}^2  - (c_t Z_{1:t-1})^2  - c_{t-1}Z_{1:t-1}^2 \right]\\
		& = \E \left[  (c_t + c_t^2) Z_{1:t-1}^2  - c_{t-1}Z_{1:t-1}^2 \right]\\
		& = \E \left[  c_{t-1} Z_{1:t-1}^2  - c_{t-1}Z_{1:t-1}^2 \right]  \quad =   \quad 0.\\
	\end{align*}
	Hence, $Q_t = Q_{t-1}$ and we are done.
\end{proof}


\section{A Few More Results on $\Phi$}

The following result is in \cite{FundConvAnalysis},  Theorem 3.3.1.
\begin{theorem}
	Let $S_1$ and $S_2$ be nonempty closed convex sets and $\sigma_1$, $\sigma_2$ are their respective support functions. Then
	$$S_1\subset S_2 \ \ \ \Leftrightarrow \ \ \ \ \sigma_1(x) \leq \sigma_2 (x) \ \ \ \mbox{ for all } x\in \R^d $$
\end{theorem}
Hence, taking a subset of $\co \left[ -\ell(\F) \right]$ leads to a lower bound on the support function and, hence, on $\Phi$. We have an obvious corollary:

\begin{corollary}
	Suppose $S_1 \subset S \subset S_2$, where $S=\co \left[ -\ell(\F)] \right]$ and all the sets are nonempty, closed and convex. Then
	$$\Phi_1 \leq \Phi \leq \Phi_2,$$
	where $\Phi_{i}(p) = \min_{-\ell_f\in S_i} \la -\ell_f, p \ra$.  
\end{corollary} 
We remark that argmin of $\Phi_i$ can now be a loss vector of a function not found in $\F$. However, this possibility can be eliminated if $S_i$ are constructed as convex hulls of a subset of $-\ell(\F)$.

Let us now consider linear transformations of the loss class.
\begin{proposition}[Proposition 3.3.3 in \cite{FundConvAnalysis}]
	Let $A:\R^n\to\R^m$ be a linear operator, with adjoint $A^*$ (for some scalar product $\la\la \cdot, \cdot, \ra\ra$ in $\R^m$). For $S\subset \R^n$ nonempty, we have
	$$ \sigma_{\text{cl} A(S)}(y) = \sigma_S (A^*y) \ \ \ \ \text{for all } y\in \R^m$$
\end{proposition}

We can now study linear transformations of the set $\co \left[ -\ell(\F) \right]$. Suppose $A$ is a linear invertible transformation and assume the set $S$ contains a non-singular exposed face, implying a $\sqrt{T}$ rate. If the transformation $A$ is such that the set $A(S)$ still contains the exposed face (i.e. does not rotate it away from the origin), then the minimax regret over the modified $\Phi$ is also $\sqrt{T}$. Moreover, it should be differing from the original regret by a property of $A$, such as the condition number. We can use the idea of transformations $A$ to define {\em isomorphic} learning problems, i.e. those which can be obtained by some invertible mapping of the loss class.

\bibliographystyle{plain}
\bibliography{regret}

\section*{Appendix} 

\begin{proof}[Proof of Theorem~\ref{theorem:maximin} for general $T$]
	Consider the last optimization choice $z_T$ in
	Eq. \eqref{eq:def_minimax_regret}. Suppose we instead draw $z_T$ according to a distribution, and compute the expected value of the quantity in the parentheses in Eq.~\eqref{eq:def_minimax_regret}. Then it is clear that maximizing this expected value over all distributions on $\Z$ is equivalent to maximizing over $z_T$, with the optimizing distribution concentrated on the optimal point. 
	Hence,
	\begin{align}  
		\label{eq:regret_start2}
	    \Reg_T = \inf_{f_1\in \F}\sup_{z_1\in\Z} \cdots  \inf_{f_{T-1}\in \F}\sup_{z_{T-1}\in\Z} \inf_{f_T\in \F}\sup_{p_T \in \Probs} \E_{Z_T\sim p_T}\left[ \sum_{t=1}^T\ell(z_t,f_t) - \inf_{f\in \F}\sum_{t=1}^T \ell(z_t,f)\right]. 
	\end{align}
	In the last expression, it is understood that sums are over the sequence $\{z_1,\ldots, z_{T-1}, Z_T\}$. The first $T-1$ elements are quantified in the suprema, while the last $Z_T$ is a random variable. Let us adopt the following notation for the conditional expectation: $\E_t [X] = \E_{Z_t\sim p_t}[X| z_1^{t-1}]$.

	We now apply Proposition \ref{prop:vonneumann} to the last $\inf/\sup$ pair in \eqref{eq:regret_start2} with 
	$$M(f_T,p_T) = \E_T\left[ \sum_{t=1}^T \ell(z_t, f_t) - \inf_{f\in\F}\sum_{t=1}^T \ell(z_t, f)\right],$$
	which is convex in $f_T$ (by assumption) and linear in $p_T$. Moreover, the set $\F$ is compact, and both $\F$ and $\Probs$ are convex. We conclude that
	\begin{align}
		\label{eq:one_swap}
	    \Reg_T &= \inf_{f_1\in \F}\sup_{z_1\in\Z} \cdots \inf_{f_{T-1}\in \F}\sup_{z_{T-1}\in\Z}\sup_{p_T\in \Probs} \inf_{f_T\in \F} \E_T\left[ \sum_{t=1}^T\ell(z_t,f_t) - \inf_{f\in \F}\sum_{t=1}^T \ell(z_t,f)\right] \nonumber\\
	    & = \inf_{f_1\in \F}\sup_{z_1\in\Z} \cdots \inf_{f_{T-1}\in \F}\sup_{z_{T-1}\in\Z} \sup_{p_T \in \Probs}\left( \sum_{t=1}^{T-1}\ell(z_t,f_t) +  \inf_{f_T\in \F}
	   \E_T\left[ \ell(Z_T,f_T)\right] -\E_T \inf_{f\in \F}\sum_{t=1}^T \ell(z_t,f) \right) \nonumber\\
		&= \inf_{f_1\in \F}\sup_{z_1\in\Z} \cdots \inf_{f_{T-1}\in \F}\sup_{z_{T-1}\in\Z} 
		\left[ \sum_{t=1}^{T-1}\ell(z_t,f_t) + 
		\sup_{p_T \in \Probs}
			\left\{ \inf_{f_T\in \F}
	   			\E_T\left[\ell(Z_T,f_T)\right] -\E_T \inf_{f\in \F}\sum_{t=1}^T \ell(z_t,f) 
			\right\}
			\right] \\
		&= 	\inf_{f_1\in \F}\sup_{z_1\in\Z} \cdots \inf_{f_{T-2}\in \F}\sup_{z_{T-2}\in\Z} \left(
			\sum_{t=1}^{T-2}\ell(z_t,f_t) +\right. \nonumber\\ 
			&\left.\hspace{1cm}\inf_{f_{T-1}\in \F}\sup_{z_{T-1}\in\Z} 
			\left[ \ell(z_{T-1},f_{T-1}) + 
			\sup_{p_T \in \Probs}
				\left\{ \inf_{f_T\in \F}
		   			\E_T\left[ \ell(Z_T,f_T)\right] -\E_T \inf_{f\in \F}\sum_{t=1}^T \ell(z_t,f) 
				\right\}
				\right]
				\right), \nonumber
	\end{align}
As we swap $\inf/\sup$ from inside out, $z_t$'s are taken to be random variables and denoted by $Z_t$. Below, we replace $Z_t$'s in the infima over $f_t$ of conditional expectations by a dummy variable $Z$. 
	
	It is important to note that the maximizing distribution $p_T$ depends
	on the previous choices $z_1^{T-1}$, but not on any of the $f_t$'s. 
	As before, we can replace the supremum over $z_{T-1}$ by a supremum over  distributions $p_{T-1}$. Noting that the expression inside of square brackets is convex in $f_{T-1}$ and linear in a distribution $p_{T-1}$ on $Z_{T-1}$, we invoke Proposition \ref{prop:vonneumann} again to obtain
\begin{align*} 
    \Reg_T &=
	 	\inf_{f_1\in \F}\sup_{z_1\in\Z} \cdots \inf_{f_{T-2}\in \F}\sup_{z_{T-2}\in\Z} \left(
		\sum_{t=1}^{T-2}\ell(z_t,f_t) +\right.\\ 
		&\left.\hspace{1cm}\sup_{p_{T-1}\in\Probs} \inf_{f_{T-1}\in \F} \E_{T-1}
		\left[ \ell(Z_{T-1},f_{T-1}) + 
		\sup_{p_T \in \Probs}
			\left\{ \inf_{f_T\in \F}
	   			\E_T\left[ \ell(Z,f_T) \right] -\E_T \inf_{f\in \F}\sum_{t=1}^T \ell(z_t,f) 
			\right\}
			\right]
			\right)\\
	 	&= \inf_{f_1\in \F}\sup_{z_1\in\Z} \cdots \inf_{f_{T-2}\in \F}\sup_{z_{T-2}\in\Z} 
	\left(
		\sum_{t=1}^{T-2}\ell(z_t,f_t) +\right.\\ 
		&\left.\sup_{p_{T-1}\in\Probs} 
			\left[
			\left(\inf_{f_{T-1}\in \F} \E_{T-1} \left[\ell(Z,f_{T-1})\right]\right) + 
			\E_{T-1}\sup_{p_T \in \Probs}
				\left\{ \inf_{f_T\in \F}
		   			\E_T\left[ \ell(Z,f_T) \right] -\E_T \inf_{f\in \F}\sum_{t=1}^T \ell(z_t,f) 
				\right\}
				\right]
	\right).
\end{align*}
Again, it is understood that in the last term, the sum is ranging over $\{z_1,\ldots, z_{T-2}, Z_{T-1}, Z_T\}$. Since the term in round brackets (involving $\inf_{f_{T-1}}$) does not depend on $p_T$ or $Z_{T-1}$, we can pull it inside the supremum:
\begin{align*} 
    \Reg_T 
	 	&= \inf_{f_1\in \F}\sup_{z_1\in\Z} \cdots \inf_{f_{T-2}\in \F}\sup_{z_{T-2}\in\Z} 
	\left(
		\sum_{t=1}^{T-2}\ell(z_t,f_t) +\right.\\ 
		&\left.\sup_{p_{T-1}\in\Probs} 
			\E_{T-1}\sup_{p_T \in \Probs}
				\left\{ 
					\inf_{f_{T-1}\in \F} \E_{T-1} \left[\ell(Z,f_{T-1})\right] + 
					\inf_{f_T\in \F}
		   			\E_T\left[ \ell(Z,f_T) \right] -\E_T \inf_{f\in \F}\sum_{t=1}^T \ell(z_t,f) 
				\right\}
	\right).
\end{align*}
We now argue that choosing a distribution $p_{T-1}$, averaging over the $Z_{T-1}$ under the maximizing distribution, and then maximizing over the conditional $p_T(\cdot|Z_{T-1})$ is the same as maximizing over the joint distributions $\jp_{T-1,T}$ over $(Z_{T-1},Z_T)$ and then averaging. Hence,
\begin{align*} 
    \Reg_T 
	 	&= \inf_{f_1\in \F}\sup_{z_1\in\Z} \cdots \inf_{f_{T-2}\in \F}\sup_{z_{T-2}\in\Z} 
	\left(
		\sum_{t=1}^{T-2}\ell(z_t,f_t) +\right.\\ 
		&\left.\sup_{\jp_{T-1,T}} 
			\E_{T-1}
				\left\{ 
					\inf_{f_{T-1}\in \F} \E_{T-1} \left[\ell(Z,f_{T-1})\right] + 
					\inf_{f_T\in \F}
		   			\E_T\left[ \ell(Z,f_T) \right] -\E_T \inf_{f\in \F}\sum_{t=1}^T \ell(z_t,f) 
				\right\}
	\right).
\end{align*}

Comparing this to \eqref{eq:one_swap}, we observe that the process can be repeated for the $\inf / \sup$ pair at time $(T-2)$ and so on.

\end{proof}

\begin{proof}[Proof of Lemma~\ref{lem:regret_concave}]
	Concavity of $\Phi$ is easy to establish. For any distributions $p$ and $q$, we see that
	\[
		\Phi\left(\frac{p+q}{2}\right) = \inf_{f\in\F} \E_{\frac{p+q}{2}} \ell(Z, f) \geq \frac{1}{2} \inf_{f\in\F} \E_p \ell(Z, f) + \frac{1}{2} \inf_{f\in\F} \E_q \ell(Z, f) = \frac{1}{2}\left( \Phi(p)+\Phi(q)\right).
	\]
	As for concavity of $\Reg_T$, let $p_\alpha := \alpha p + (1-\alpha)q$ and note the simple calculation of the conditional probability
	\[
		dp_\alpha(Z_t | Z_1^{t-1}) = 
		\frac{\alpha dp(Z_1^{t-1}) dp(Z_t|Z_1^{t-1}) + (1-\alpha) dq(Z_1^{t-1}) dq(Z_t|Z_1^{t-1})}
		{\alpha dp(Z_1^{t-1}) + (1-\alpha) dq(Z_1^{t-1})}.
	\]
	We can now show that
	\begin{align*}
		&\sum_{t=1}^T \E_{p_\alpha} \Phi(p_\alpha(\cdot|Z_1^{t-1})) \\
		& =  \sum_{t=1}^T  \int \Phi(p_\alpha(\cdot|Z_1^{t-1})) (\alpha dp(Z_1^{t-1}) + (1-\alpha)dq(Z_1^{t-1}))\\
		& \geq \sum_{t=1}^T 
		\int \frac{\alpha dp(Z_1^{t-1}) \Phi(p(\cdot|Z_1^{t-1})) + (1-\alpha) dq(Z_1^{t-1}) \Phi(q(\cdot|Z_1^{t-1}))}
		{\alpha dp(Z_1^{t-1}) + (1-\alpha) dq(Z_1^{t-1})}\\
		&  \quad \quad \quad \quad \quad \quad 
		\times (\alpha dp(Z_1^{t-1}) + (1-\alpha)dq(Z_1^{t-1})) \\
		& =  \sum_{t=1}^T \alpha\E_p \Phi(p(\cdot|Z_1^{t-1})) + (1-\alpha)\E_q \Phi(q(\cdot|Z_1^{t-1})).
	\end{align*}
	Thus, the first term in the regret is concave with respect to the joint distribution. In addition the second term is clearly linear, since
	\[
		- \E_{p_\alpha} \Phi(\Unif) = - \alpha \E_{p} \Phi(\Unif) - (1-\alpha)\E_{q} \Phi(\Unif).
	\]
	Since a linear plus a concave function is still concave, $\Reg_T(\cdot)$ is concave.
\end{proof}

\begin{proof}[Proof of Lemma~\ref{lem:stability}]
	 Since $\ell$ is $\sigma$-strongly convex, we have (by taking $f=f_p$, $g=f_q$ in the definition of strong convexity)
	$$\frac{\ell(z, f_p)+\ell(z, f_q)}{2} \geq \ell\left(z, \frac{f_p+f_q}{2}\right) + \frac{\sigma}{8} \|f_p-f_q\|^2
	$$
	for any $z$. Taking expectations with respect to $z\sim p$ and noting that $f_p$ minimizes $\E_p\ell(z,f)$, we have
	\begin{align*}
		\E_p\left[\frac{\ell(z, f_p)+\ell(z, f_q)}{2} \right] &\geq \E_p \ell\left(z, \frac{f_p+f_q}{2}\right) + \frac{\sigma}{8} \|f_p-f_q\|^2\\
		& \geq\E_p \ell\left(z, f_p\right) + \frac{\sigma}{8} \|f_p-f_q\|^2
	\end{align*}
	Rearranging terms,
	$$\frac{\sigma}{4} \|f_p-f_q\|^2 \leq \E_p \ell(z, f_q)- \E_p \ell(z, f_p).$$
	Similarly,
	$$\frac{\sigma}{4} \|f_p-f_q\|^2 \leq \E_q \ell(z, f_p)- \E_q \ell(z, f_q).$$
	Adding,
	$$\frac{\sigma}{2} \|f_p-f_q\|^2 \leq \int_z \left[\ell(z, f_q)-\ell(z, f_p)\right](dp(z)-dq(z)).$$
	Using the Lipschitz condition,
	\begin{align*}
		\frac{\sigma}{2} \|f_p-f_q\|^2 &\leq \int_z \left| \ell(z, f_q)-\ell(z, f_p)\right|\cdot |dp(z)-dq(z)| \\
		&\leq L\|f_p-f_q\|\cdot \|p-q\|_1
	\end{align*}
	Thus,
	\begin{align*}
		\|f_p-f_q\| \leq \frac{2L}{\sigma} \|p-q\|_1, 
	\end{align*}
	which establishes the main building block resulting from the curvature.
\end{proof}

\begin{lemma}
	If $\ell$ satisfies the conditions of Theorem~\ref{thm:flatness},
then $\ell(\cdot, f_p)$ is a subdifferential of $\Phi$ at $p$.
\end{lemma}
\begin{proof}
We claim that $\ell(\cdot, f_p)$ is the differential of $\Phi$ at the point $p$ and, therefore, $\int_z \ell(z, f_p) (dq(z)-dp(z)) = \la \nabla \Phi(p), (q-p) \ra$ is the derivative in the direction $q-p$. By definition, the differential is a function $\nabla \Phi$ such that
$$\lim_{h\to 0}\frac{\Phi(p)-\Phi(p+h)-\nabla\Phi(p)\cdot h }{\|h\|} = 0.$$
Hence, it remains to check that for any distribution $r$ 
\begin{align}
	\label{eq:derivative_def}
	\lim_{\alpha\to 0} \frac{\Phi((1 - \alpha) p+\alpha r)-\Phi(p)- \int_z \ell(z,f_p) (\alpha (dr(z) - dp(z)))} {\alpha} =0.
\end{align}
Rewriting,
\begin{align*}  
        &\Phi((1 - \alpha) p+\alpha r)-\Phi(p)- \int_z \ell(z,f_p) (\alpha (dr(z) - dp(z))) \\
	&= \min_f \E_{(1-\alpha)p+\alpha r}\ell(z,f) - (1-\alpha)\min_f \E_{p} \ell(z,f) - \alpha\E_r \ell(z, f_p) \\
	&= \min_f \left[(1-\alpha)\E_p \ell(z,f)+\alpha\E_r \ell(z,f)\right] - (1-\alpha)\E_{p} \ell(z,f_p) - \alpha\E_r \ell(z, f_p)
\end{align*}
It is evident that the above expression is non-positive by substituting a particular choice of $f_p$ in the first minimum. For the lower bound, use the bound of Eq. \eqref{eq:curvature}
\begin{align*}  
	&\E_{(1-\alpha)p+\alpha r}\ell(z,f_{(1-\alpha)p+\alpha r}) - (1-\alpha)\E_{p} \ell(z,f_p) - \alpha\E_r \ell(z, f_p)\\ 
	&=\E_{(1-\alpha)p+\alpha r}\ell(z,f_{(1-\alpha)p+\alpha r}) - \E_{(1-\alpha)p+\alpha r}\ell(z,f_p) \\
	&\geq -\frac{2L^2}{\sigma} \|\alpha(p - r)\|_1^2 = -\Theta(\alpha^2)
\end{align*}
Thus, Eq. \eqref{eq:derivative_def} is verified.
\end{proof}

\end{document}

%% file: macros.tex
\newtheorem{proposition}[theorem]{Proposition}

\newcommand{\Rad}{\ensuremath{\widehat{\mbox{Rad}}}}
\newcommand{\la}{\langle}
\newcommand{\ra}{\rangle}
\newcommand{\tr}{\ensuremath{{\scriptscriptstyle\mathsf{T}}}} 
\newcommand\E{{\mathbb E}\,}
\newcommand\R{{\mathbb R}}
\newcommand\F{{\mathcal F}}
\newcommand\Z{\mathcal Z}
\newcommand\Reg{\mathscr R}
\newcommand\co{\text{co}}
\newcommand\Probs{\mathscr P}
\newcommand{\Tinv}{\ensuremath{\frac{1}{T}}}

\newcommand{\Dvg}{\ensuremath{\mathcal D}}
\newcommand{\Unif}[1][T]{\ensuremath{\hat P_{#1}}}
\newcommand{\Ub}[1][T]{\ensuremath{\bar{P}_{#1}}}
\newcommand{\pt}{\ensuremath{p_t\left(\cdot|Z_1^{t-1}\right)}}
\newcommand{\ptm}{\ensuremath{p_t^m}}
\newcommand{\ptmp}{\ensuremath{p_{t'}^m}}

\newcommand{\jp}{\ensuremath{\mathbf{p}}}

\newcommand{\one}{\mathbf{1}}

\newcommand{\frameitblack}[1]{
    \begin{center}
    \framebox[.95\columnwidth][l]{
        \begin{minipage}{.9\columnwidth}
        #1
        \end{minipage}
    }
    \end{center}
}